\documentclass[final]{cvpr}

\usepackage{times}
\usepackage{epsfig}
\usepackage{graphicx}
\usepackage{amsmath}
\usepackage{amssymb}
\usepackage{amsfonts}
\usepackage{subfig}
\usepackage[american]{babel}
\usepackage{url}
\usepackage{xcolor}
\usepackage{graphicx}
\usepackage{multirow}
\usepackage{booktabs} 
\usepackage{pifont}
\newcommand{\cmark}{\ding{51}}%
\newcommand{\xmark}{\ding{55}}%
\usepackage[misc]{ifsym}
\usepackage{dblfloatfix}
\usepackage{transparent}
\usepackage{algorithm}
\usepackage{listings}

\newcommand{\app}{\raise.17ex\hbox{$\scriptstyle\sim$}}

\definecolor{Highlight}{HTML}{39b54a}  
\newcommand{\hl}[1]{\textcolor{Highlight}{#1}}
\makeatletter\renewcommand\paragraph{\@startsection{paragraph}{4}{\z@}
  {.5em \@plus1ex \@minus.2ex}{-.5em}{\normalfont\normalsize\bfseries}}\makeatother

\usepackage{etoolbox}
\makeatletter
\AfterEndEnvironment{algorithm}{\let\@algcomment\relax}
\AtEndEnvironment{algorithm}{\kern2pt\hrule\relax\vskip3pt\@algcomment}
\let\@algcomment\relax
\newcommand\algcomment[1]{\def\@algcomment{\footnotesize#1}}
\renewcommand\fs@ruled{\def\@fs@cfont{\bfseries}\let\@fs@capt\floatc@ruled
  \def\@fs@pre{\hrule height.8pt depth0pt \kern2pt}%
  \def\@fs@post{}%
  \def\@fs@mid{\kern2pt\hrule\kern2pt}%
  \let\@fs@iftopcapt\iftrue}
\makeatother

\usepackage[pagebackref=true,breaklinks=true,colorlinks,bookmarks=false]{hyperref}
\def\cvprPaperID{} 
\def\confYear{CVPR 2021}

\begin{document}

\title{Batch Normalization with Enhanced Linear Transformation}

\author{%
Yuhui Xu\textsuperscript{1}\footnotemark[1] \quad
Lingxi Xie\textsuperscript{2} \quad
Cihang Xie\textsuperscript{3} \quad
Jieru Mei\textsuperscript{2} \quad 
Siyuan Qiao\textsuperscript{2} \quad 
Wei Shen\textsuperscript{1} \quad\\ 
Hongkai Xiong\textsuperscript{1}\quad
Alan Yuille\textsuperscript{2} \vspace{.3em}\\
\!\!\!\textsuperscript{1}Shanghai Jiao Tong University \quad 
\textsuperscript{2}Johns Hopkins University \quad
\textsuperscript{3}University of California, Santa Cruz
\vspace{-0.35em}
}

\maketitle
 \renewcommand*{\thefootnote}{\fnsymbol{footnote}}
 \setcounter{footnote}{1}
 \footnotetext{This work was done when Y. Xu was a visiting student at JHU.}
 \renewcommand*{\thefootnote}{\arabic{footnote}}
 \renewcommand*{\thefootnote}{\fnsymbol{footnote}}

\begin{abstract}
Batch normalization (BN) is a fundamental unit in modern deep networks, in which a linear transformation module was designed for improving BN's flexibility of fitting complex data distributions. In this paper, we demonstrate properly enhancing this linear transformation module can effectively improve the ability of BN. Specifically, rather than using a single neuron, we propose to additionally consider each neuron's neighborhood for calculating the outputs of the linear transformation.
Our method, named BNET, can be implemented with 2--3 lines of code in most deep learning libraries. Despite the simplicity, BNET brings consistent performance gains over a wide range of backbones and visual benchmarks. Moreover, we verify that BNET accelerates the convergence of network training and enhances spatial information by assigning the important neurons with larger weights accordingly. The code is available at \url{https://github.com/yuhuixu1993/BNET}.
\end{abstract}

\section{Introduction}\label{Sec:Intro}

Deep learning has reshaped the computer vision community in recent years~\cite{lecun2015deep}. Most vision problems, including classification~\cite{simonyan2014very,he2016deep},  detection~\cite{ren2015faster,lin2017feature}, and  segmentation~\cite{chen2017deeplab,zhao2017pyramid}, fall into the pipeline that starts with extracting image features using deep networks. When training very deep networks, batch normalization (BN)~\cite{ioffe2015batch} is a standard tool to regularize the distribution of neural responses so as to improve the numerical stability of optimization.

There are many variants of BN, differing from each other mostly in the ways of partitioning input data, including by instances~\cite{ulyanov2016instance}, channels~\cite{pan2018two}, groups~\cite{wu2018group}, positions~\cite{li2019positional}, and image domains~\cite{zhuang2020rethinking}. These variants generally shared the same module, often referred to as linear transformation, which applies a pair of learnable coefficients to restore the representations of the normalized neural responses. As a result, the output is no longer constrained within a zero-mean, unit-variance distribution, and the model has a stronger ability in fitting the real data distribution.

\begin{table}[t]
\small
    \centering
    \setlength{\tabcolsep}{0.12cm}
    \begin{tabular}{l|l|lcc}
    \toprule
        Task & Norm & Performance & \#Params & GFLOPs\\
        \midrule
        \multirow{2}*{ImageNet~\cite{deng2009imagenet}} & BN & \multicolumn{1}{l}{76.3} & 25.6 & 4.1\\
                          & BNET-$3$ & 76.8~\hl{(${+}$\textbf{0.5})} & 25.7 & 4.2 \\
        \midrule
        \multirow{2}*{MS-COCO~\cite{Lin2014MicrosoftCC}} & BN & \multicolumn{1}{l}{37.5} & 41.5 & 207.1\\
                          & BNET-$3$ & 39.5~\hl{(${+}$\textbf{2.0})} & 41.6 & 208.1 \\
        \midrule
        \multirow{2}*{Cityscapes~\cite{cordts2016cityscapes}} & BN & \multicolumn{1}{l}{76.3} & 49.0 & 354.4\\
                          & BNET-$3$ & 77.4~\hl{(${+}$\textbf{1.1})} & 49.1 & 355.8 \\
        \midrule
        \multirow{2}*{UCF-101~\cite{soomro2012ucf101}} & BN & \multicolumn{1}{l}{79.8} & 23.7 & 4.3\\
                          & BNET-$3$ & 84.1~\hl{(${+}$\textbf{4.3})} & 23.8 & 4.4 \\
    \bottomrule
    \end{tabular}
    \caption{Compared to BN, BNET-$3$ helps ResNet-50~\cite{he2016deep} gain consistent improvements on four different visual tasks, including classification, detection, segmentation and action recognition. The corresponding evaluation metrics are accuracy, AP, mIOU, and accuracy, respectively.
    Note that the extra computational costs added by BNET are negligible.}
    \vspace{-0.2em}
    \label{tab:summaryresults}
\end{table}

The above analysis implies that the learning ability of the linear transformation module in BN affects the flexibility of deep networks. However, we notice that existing methods have mostly assumed the linear transformation module takes a single neuron as the input and outputs the corresponding response---the referable information generally is much fewer than other operations like convolution and pooling. Such a design of the linear transformation module potentially limit the ability of BN in fitting much more complex data distributions.

To this end, we present a straightforward method to enhance the standard BN---rather than just use a single neuron, we allow the linear transformation module to calculate the output based on a set of neighboring neurons in the same channel. Our method is easy to implement, \eg, as shown in Algorithm \ref{alg:pytorch_BNET}, PyTorch \cite{paszke2019pytorch} instantiation can be as simple as 
switching off the \texttt{affine} option in BN and appending a channel-wise convolution~\cite{chollet2017xception}  afterwards.
This logic is easily transplanted to other deep learning libraries and can be implemented within 2--3 lines of code. We name this modified BN as BNET, short for \textbf{BN with Enhanced Transformation}, and use BNET-$k$ to indicate a $k\times k$ channel-wise convolution being used for linear transformation.

We demonstrate the effectiveness of the proposed BNET on a wide range of network backbones in several visual benchmarks. As shown in Table~\ref{tab:summaryresults}, by simply replacing BN with BNET, the standard ResNet-50 backbone achieves a $76.8\%$ (+$0.5\%$) top-1 accuracy on ImageNet classification, a $39.5\%$ (+$2.0\%$) AP on MS-COCO detection, and a $77.4\%$ (+$1.1\%$) mIOU on Cityscapes semantic segmentation. BNET works better if a large kernel size is adopted, \eg, compared to BNET-$3$, BNET-$7$ can further improve the detection AP by $1.2\%$ on MS-COCO at marginally increased costs of $\app2\%$ more FLOPs. More results are provided in the experimental section.

Besides these quantitative results, we also show that BNET enjoys faster convergence in network training.
This property saves computational costs especially
when strong regularizations (\eg, AutoAugment~\cite{cubuk2019autoaugment}) are added to the training framework---by equipping networks with BNET, the demands of requiring extra training epochs is much alleviated. 
Additionally, by performing linear regression on the input-output pairs of BNET, an interesting observation is that the neurons that are related to important objects or contexts will be significantly enhanced. 
This observation offers a conjecture that the faster convergence property of BNET is partly due to its stronger ability in capturing spatial cues.

In summary, the main contribution of this paper lies in a simple enhancement of BN that consistently improves recognition models. Our study delivers an important messages to the community that batch normalization needs a tradeoff between normalization and flexibility. We look forward to more products along this research direction.

\section{Related Works}\label{Sec:Rela}

\paragraph{Normalization methods.} Layer Response Normalization~(LRN) was an early normalization method that computed the statistics in a small neighborhood of each pixel and was applied in early models~\cite{hinton2012imagenet}. Batch Normalization~(BN)~\cite{ioffe2015batch} accelerated training and improved generalization by computing the mean and variance more globally along the batch dimension. By contrast, Layer Normalization~(LN)~\cite{ba2016layer} computed the statistics of all channels in a layer and was proven to help the training of recurrent neural networks. Instance Normalization~(IN)~\cite{ulyanov2016instance} performed normalization along each individual channel and was commonly used in image generation~\cite{huang2017arbitrary}. A pair of parameters $\gamma$ and $\beta$ were learned to scale and shift the normalized features along the channels. The above three normalization methods can also be applied jointly, $e.g.$, IBN-Net~\cite{pan2018two} integrated IN and BN to eliminate the appearance in DNNs and Switchable Normalization~(SN)~\cite{luo2018differentiable} combines all the three normalizers using a differentiable method to learn the ratio of each one. Instead of normalizing features, Weight Normalization~(WN)~\cite{salimans2016weight} and Weight Standardization~(WS)~\cite{qiao2019weight} proposed to normalize the weights. Decorrelated Batch Normalization~(DBN)~\cite{huang2018decorrelated} extended BN by decorrelating features using the covariance matrix computed over a mini-batch. However, most of the proposed normalization methods focused ``where'' and ``how'' to normalize and ignore the linear recovery part of these normalization methods.

\paragraph{Understanding the normalization in DNNs.} Many efforts have been made to understand and analyze the effectiveness of BN. Cai~\etal~\cite{cai2019quantitative} proved the convergence of gradient descent with BN (BNGD) for arbitrary learning rates for the weights. Santurkar~\etal~\cite{santurkar2018does} found that BN made the optimization landscape significantly smoother. Bjorck~\etal~\cite{bjorck2018understanding} empirically showed that BN enabled training with large gradients steps which may result in diverging loss and activations growing uncontrollably with network depth. Li~\etal~\cite{li2019understanding} found a way which can jointly use dropout~\cite{srivastava2014dropout} and BN to boost the performance. In addition to directly analyzing the effectiveness of BN, some works try to train DNNs without BN to better understand the normalization layers. Zhang~\etal~\cite{zhang2019fixup} introduced a new initialization method to solve the exploding and vanishing gradient problem without BN.  Qi~\etal~\cite{qi2020deep} successfully trained deep vanilla ConvNets without normalization nor skip connections by enforcing the convolution kernels to be near isometric during initialization and training. Shao~\etal~\cite{shao2020normalization} proposed RescaleNet which can be trained without normalization layers and without performance degradation. \emph{In this paper, we find that the recovery parameters are important for the training of DNNs and improve it by incorporating contextual information.}

\section{Method}\label{Sec:Method}
\subsection{Background and Motivations}
Batch Normalization is proposed to normalize each channel of input features into zero mean and unit variance. Considering the input of a layer in DNNs over a mini-batch: the inputs share the same channel index of $N$ samples are normalized together.
\begin{equation}
    \hat{\mathbf{x}}^{(n)}_c=\frac{1}{\sigma_c}(\mathbf{x}^{(n)}_c-\mu_c),\label{equ:bn}
\end{equation}
where $c$ and $n$ are the channel index and sample index, respectively.
$\mu$ and $\sigma$ in Equation~\eqref{equ:bn} are the mean and standard deviation computed as follows:
\begin{equation}
    \mu_c=\frac{1}{N}\sum_{n=1}^N \mathbf{x}^{(n)}_c, \sigma_c = \sqrt{\frac{1}{N}\sum_{n=1}^N (\mathbf{x}^{(n)}_c-\mu_c)^2+\epsilon},\label{equ:meanstd}
\end{equation}
where $\epsilon$ is a small constant. In order to recover the representation ability of the normalized feature, a pair of per-channel parameters $\gamma$ and $\beta$ is learned.
\begin{equation}
    \mathbf{y}^{(n)}_c=\gamma\hat{\mathbf{x}}^{(n)}_c+\beta\label{bnrecover}
\end{equation}

\begin{figure}[t]
	\renewcommand{\baselinestretch}{1.0}
	\centering
	\subfloat[BN]{
		\includegraphics[width=0.20\textwidth]{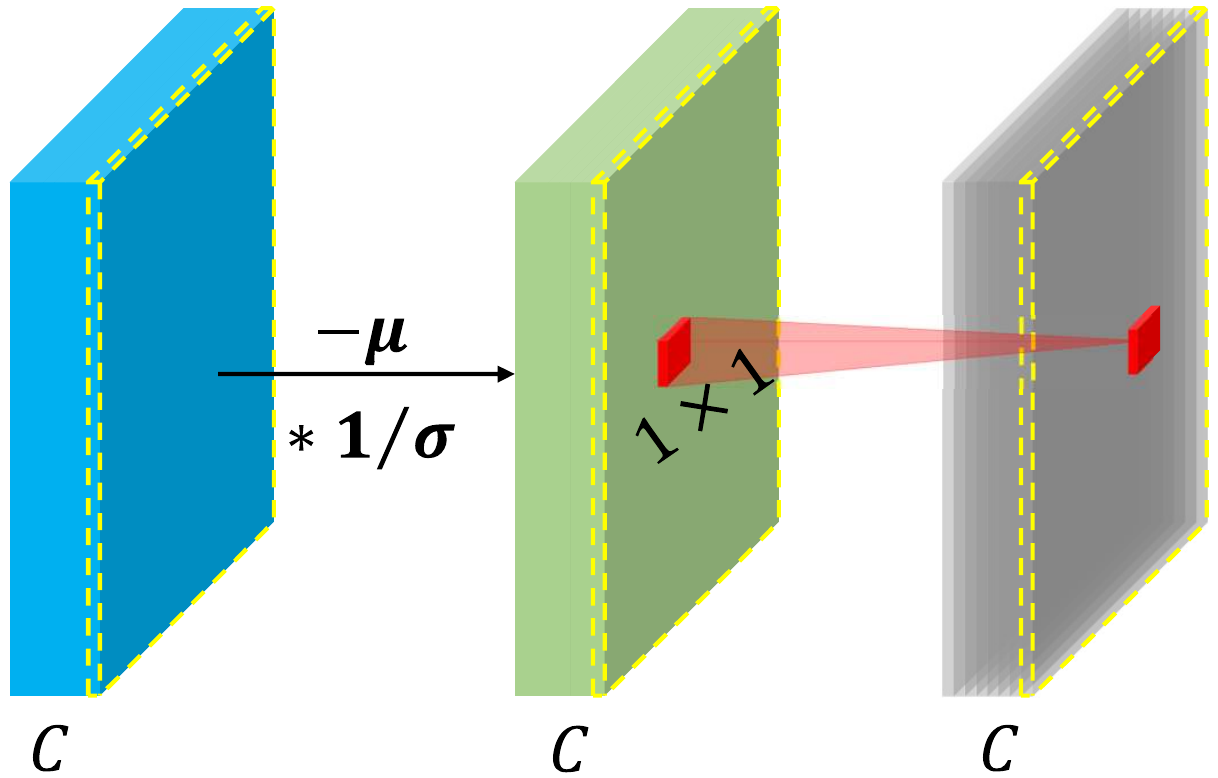}\label{fig.BN}}\hspace{5mm}
	\subfloat[BNET-$k$]{
		\includegraphics[width=0.20\textwidth]{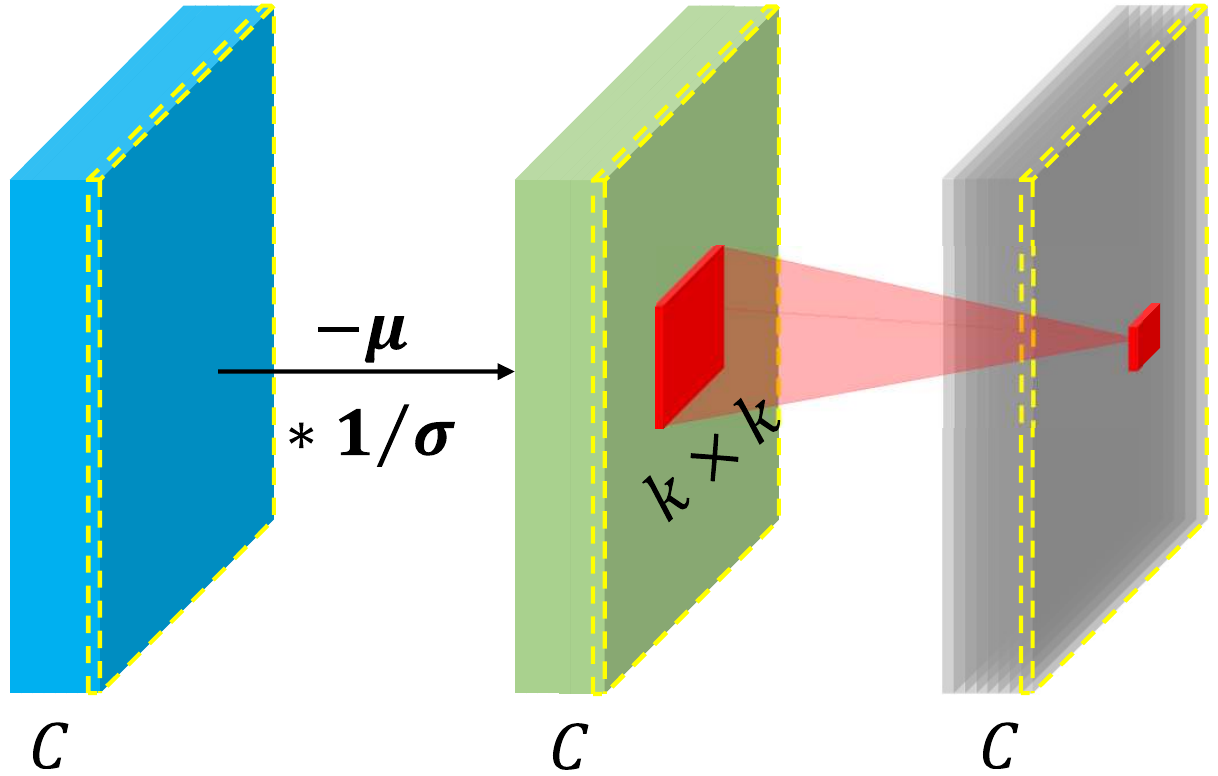}\label{fig.BNET}}
	\caption{The framework of BN and BNET-$k$. We consider the per-channel parameters $\{\gamma,\beta$\} as an equivalent $1\times 1$ \textit{depth-wise convolution} to help better understanding the extended BNET-$k$.}\label{fig.framework}
\end{figure}

Contextual information has been verified important in many visual benchmarks. Contextual conditioned activations~\cite{goodfellow2013maxout,ma2020funnel} were proposed to improve the performance of visual recognition. SE-Net utilized attention modules to provide contextual information. Non-local networks~\cite{wang2018non} captured the long-range dependencies and facilitate many tasks. Relation networks~\cite{hu2018relation} modeled the relation between objects by means of their appearance feature and geometry to improve object detection. One important role of BN is fitting the distribution of the inputs by recovering the statistics of the inputs. The previous linear recovery assumes that the input pixels are spatial-wise independent which is the opposite of the case. For the input with a complex environment and multiple objects, it is hard to capture its distribution with the independence assumptions. In such cases, contextual information is very important. In this paper, we mainly focus on how to embed contextual information in BN. For a neighborhood $\mathcal{S}_{\hat{x}_i}$ of the normalized input $\hat{x}_i$, we give a generalized formulation of the recovery step in BN which considers the contextual information and local dependencies:
\begin{equation}
    y_i=\sum_{j\in \mathcal{S}_{\hat{x}_i}}f_j(\hat{x}_j)+\beta,\label{general}
\end{equation}
where $j$ is the index of the neighbors and $f_j$ is the recover functions. BN is a special case of Equation~\eqref{general} when only $\hat{x}_i$ itself is included in ${S}_{\hat{x}_i}$.

\subsection{The Formulation of BNET}
Different from previous works that mainly focus on studying different ways to calculate normalization statistics, we hereby investigate the effects of the linear recovery transformations of BN. The primitive target of linear transformation parameters $\gamma$ and $\beta$ is to fit the distribution of the input and recover representation ability, however, two simple parameters of each channel can hardly accomplish this job without contextual information provided especially when the input contains complex scenes.

Here, based on the general formulation in Equation~\eqref{general}, we adopt a parameterized linear transformation $M{(\cdot;\boldsymbol{\theta})}$ for simplicity. Therefore, the recovered features can be computed as follows: 
\begin{equation}
    \mathbf{y}^{(n)}_c=M{(\hat{\mathbf{x}}^{(n)}_c;\boldsymbol{\theta})}+\beta,\label{BNET}
\end{equation}
where $\boldsymbol{\theta}$ denotes the learned parameters.

To capture contextual information, a simple instantiation of $M{(\cdot;\boldsymbol{\theta})}$ is by using $(k\times k)$ \textit{depth-wise convolution}~\cite{chollet2017xception}. We name this method as BNET-$k$, short for Batch Normalization with Enhanced Linear Transformation, where $k$ denotes the kernel size of depth-wise convolution.
An illustration is provided in Figure~\ref{fig.framework}. BNET-$k$ can be easily implemented by a few lines of code based on the original implementation of BN in PyTorch~\cite{paszke2019pytorch} and TensorFlow~\cite{abadi2016tensorflow}. Algorithm~\ref{alg:BNET} provides the code of BNET in PyTorch.
\begin{algorithm}[t]
\caption{BNET code in PyTorch-like style}\label{alg:pytorch_BNET}
\label{alg:BNET}
\definecolor{codeblue}{rgb}{0.25,0.5,0.5}
\lstset{
  backgroundcolor=\color{white},
  basicstyle=\fontsize{7.5pt}{7.5pt}\ttfamily\selectfont,
  columns=fullflexible,
  breaklines=true,
  captionpos=b,
  commentstyle=\fontsize{8pt}{8pt}\color{codeblue},
  keywordstyle=\fontsize{7.2pt}{7.2pt},
}
\begin{lstlisting}[language=python]
# width: number of input channels
# k: the kernel size of the transformation

class BNET2d(nn.BatchNorm2d):

   def __init__(self, width, *args, k=3, **kwargs):
      super(BNET2d, self).__init__(width, *args, affine=False, **kwargs)
      self.bnconv = nn.Conv2d(width, width, k, padding=(k-1) // 2, groups=width, bias=True)

   def forward(self, x):
      return self.bnconv(super(BNET2d, self).forward(x))
\end{lstlisting}
\end{algorithm}
\subsection{Advantages over Previous Mechanisms}

BNET offers a plug-and-play option to improve BN, a standard module in most modern deep networks. Before continuing with extensive experiments, we briefly analyze its advantages over some past works.

To the best of our knowledge, this is the first work that improves BN by enhancing the linear transformation. It is complementary to prior BN variants that mainly contributed to partitioning the input data into different groups~\cite{ba2016layer,ulyanov2016instance,wu2018group}. BNET requires additional parameters to enhance the linear transformation, but the increased amount is often much smaller than that of the convolutional parameters. In other words, we find an objective that has been undervalued and verify that small extra costs lead to big gains.

Another work that relates to BNET is FReLU~\cite{ma2020funnel}, which introduced context information to the activation function (\textit{e.g.}, ReLU). We point out that BNET enjoys a stronger ability in fitting different distributions, and BNET can be combined with FReLU for better recognition performance. In ImageNet classification on ResNet-50, adding BNET-$3$ upon FReLU improves the top-1 accuracy from 77.5\% to 77.9\%, and the benefit transfers to MS-COCO object detection, claiming an AP gain of 0.4\% (from 39.8\% to 40.2\%).


\section{Experiments}\label{Sec:Exp}
In this section, we will present the experimental results, including image classification on ImageNet~\cite{deng2009imagenet}, object detection and instance segmentation on MS-COCO~\cite{Lin2014MicrosoftCC}, video
recognition on UCF-101~\cite{soomro2012ucf101} and semantic segmentation on Cityscapes~\cite{cordts2016cityscapes}.

\subsection{Image Classification}
To better evaluate the effectiveness of our proposed BNET, we conduct image classification experiments on ImageNet~\cite{deng2009imagenet}. It contains $1\rm{,}000$ object categories, and $1.3\mathrm{M}$ training images and $50\mathrm{K}$ validation images, all of which are high-resolution and roughly equally distributed over all classes. Various architectures are adopted. First, we apply BNET on ResNet~\cite{he2016deep} and ResNeXt~\cite{xie2017aggregated}. Then, experiments on efficient models~(MobileNetV2~\cite{Sandler2018MobileNetV2IR}) and low-precision models~(Bi-real Net~\cite{liu2018bi}) are presented. Experiments of training with  Auto-Augment~\cite{cubuk2019autoaugment} are discussed, afterwards. In these experiments, the input image size is fixed to be $224\times224$. 
\subsubsection{Comparison on ResNet and ResNeXt}\label{Sec:resnet}
For a fair comparison, we train all the models in the same code base with the same settings. The learning rate is set to $0.1$ initially, and is multiplied by $0.1$ after every 30 epochs. We use SGD to train the models for a total of 100 epochs, where the weight decay is set to 0.0001 and the momentum is set to 0.9. For ResNet-18 and ResNet-50, the training batch is set to 256 for 4 GPUs~(Nvidia 1080Ti). We use 8 GPUs~(Nvidia 1080Ti) to train ResNet-101 and ResNeXt-50 with a batch-size 256.

\begin{figure*}[!t]
	\renewcommand{\baselinestretch}{1.0}
	\centering
	\subfloat[Loss w/o Auto-Augment]{
		\includegraphics[width=0.24\textwidth]{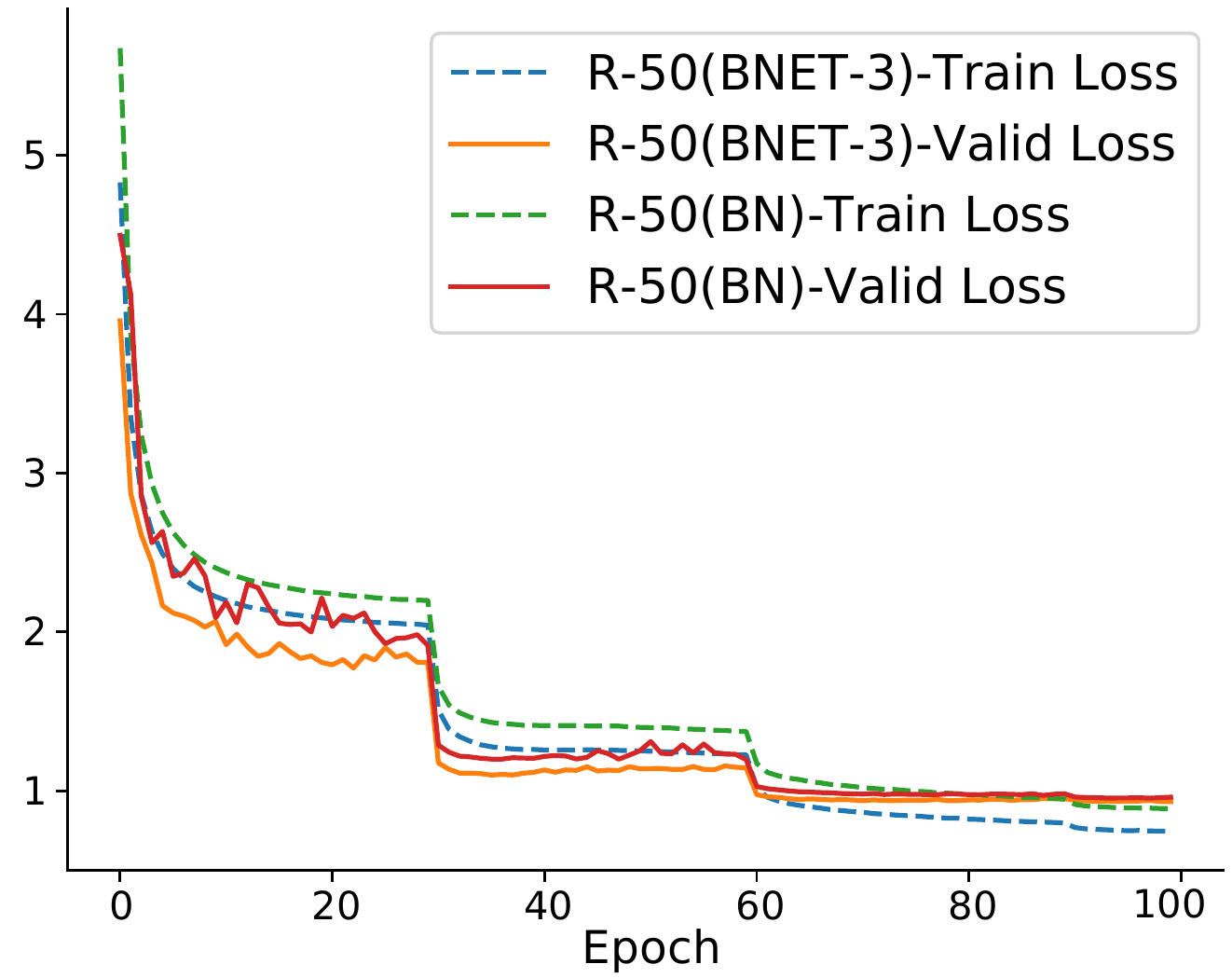}\label{fig.r50loss}}
	\subfloat[Acc. w/o Auto-Augment]{
		\includegraphics[width=0.24\textwidth]{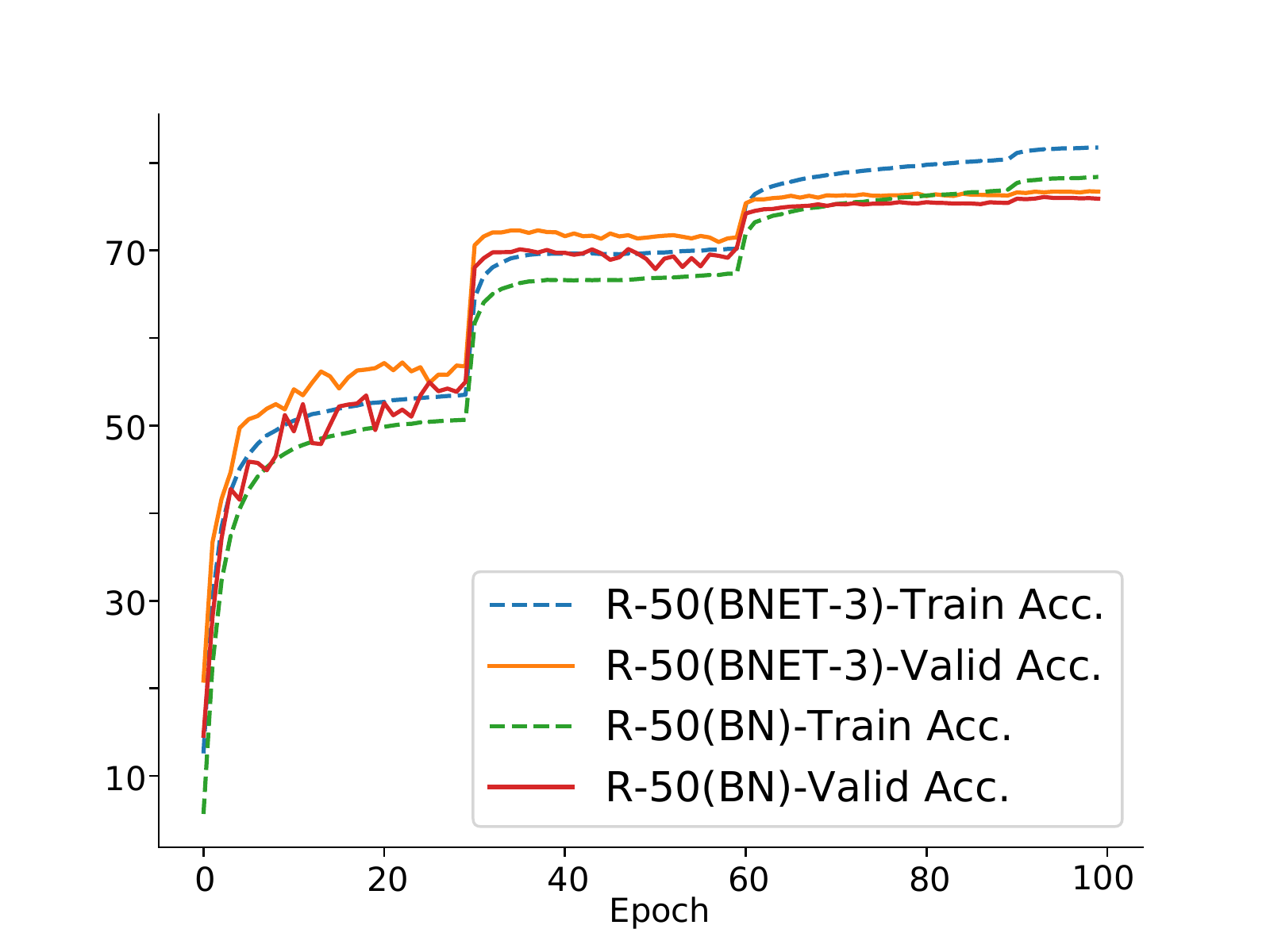}\label{fig.r50acc}}
	\subfloat[Loss with Auto-Augment]{
		\includegraphics[width=0.24\textwidth]{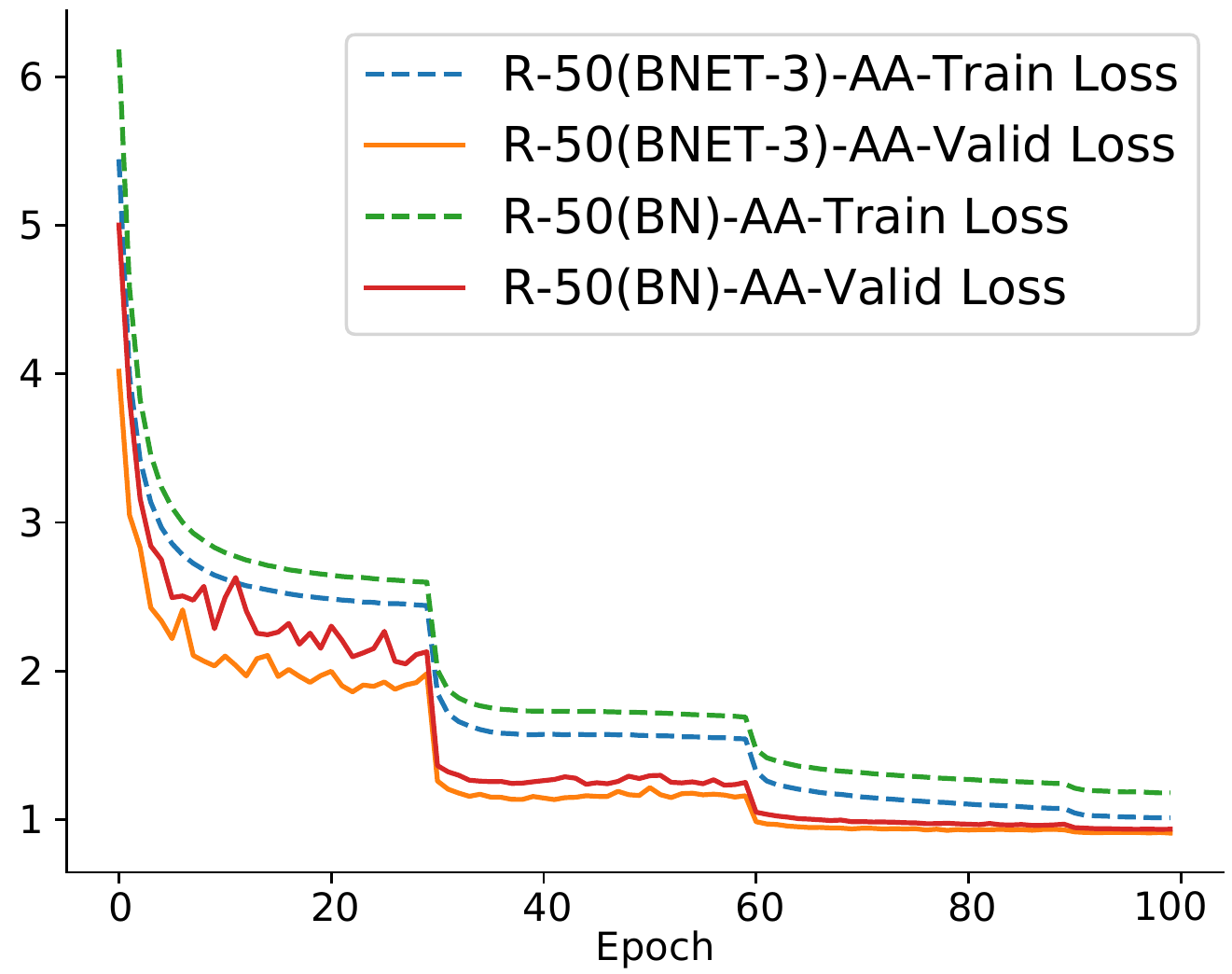}\label{fig.r50lossauto}}
	\subfloat[Acc. with Auto-Augment]{
		\includegraphics[width=0.24\textwidth]{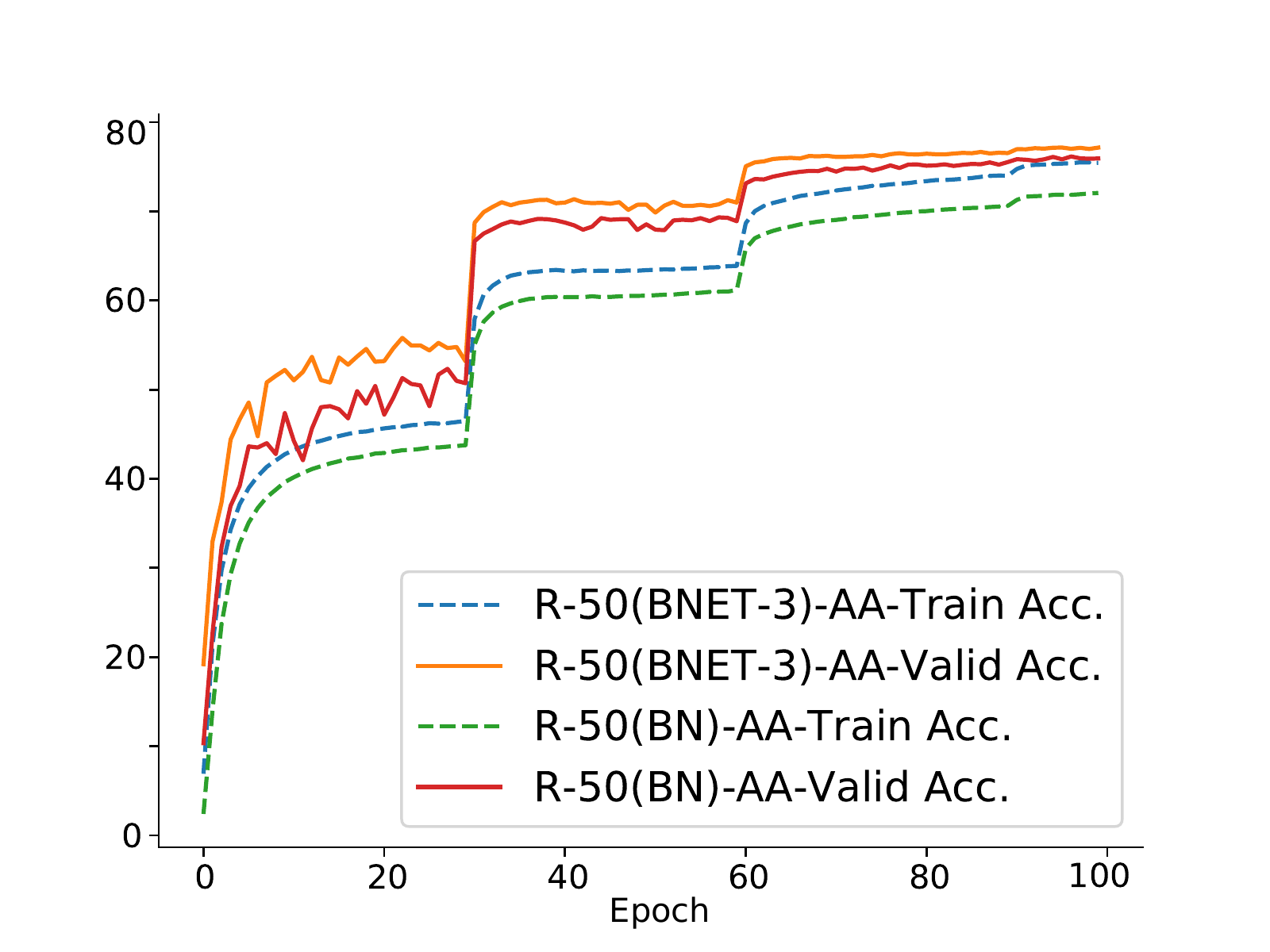}\label{fig.r50accauto}}
	\caption{The training curves of ResNet-50 using BN and BNET-$3$ on ImageNet. We consider two situations: Figure~\ref{fig.r50loss} and Figure~\ref{fig.r50acc} are the loss and accuracy curves without using Auto-Augment~\cite{cubuk2019autoaugment} while Figure~\ref{fig.r50lossauto} and Figure~\ref{fig.r50accauto} are the loss and accuracy curves using Auto-Augment.}\label{fig.R50}
\end{figure*}

Table~\ref{tab:resnet} shows the major experimental results on ImageNet. The proposed BNET consistently outperforms BN on ResNet with different depths and ResNetXt-50 with negligible computation cost. For example, ResNet-101 with BNET-$3$ has a remarkable increase of 0.9\% on Top-1 and 0.4\% on Top-5. Figure~\ref{fig.r50loss} and Figure~\ref{fig.r50acc} depict the loss and accuracy curves of ResNet-50~(BN) and ResNet-50~(BN-$3$). We find that model using BNET converges faster than the model using BN.  
\begin{table}[t]
\small
	\resizebox{\linewidth}{!}{
    \centering
    \begin{tabular}{l|l|c|c|c|c}
    \toprule
        Model & Norm & \#Params & GFLOPs & Top-1&Top-5\\
        \midrule
        R-18 & BN & 11.7 & 1.8 & \multicolumn{1}{l|}{69.6} & \multicolumn{1}{l}{89.1}\\
        R-18 & BNET-$3$ & 11.7 & 1.8 & 70.3~\hl{(${+}$\textbf{0.7})} & 89.4~\hl{(${+}$\textbf{0.3})}\\
        \midrule
        R-50 & BN & 25.6 & 4.1 & \multicolumn{1}{l|}{76.3} & \multicolumn{1}{l}{93.0}\\
        R-50 & BNET-$3$ & 25.7 & 4.2 & 76.8~\hl{(${+}$\textbf{0.5})} & 93.1~\hl{(${+}$\textbf{0.1})}\\
        \midrule
        R-101 & BN & 44.5 & 7.8 & \multicolumn{1}{l|}{77.4} & \multicolumn{1}{l}{93.5}\\
        R-101 & BNET-$3$ & 44.8 & 7.9 & 78.3~\hl{(${+}$\textbf{0.9})} & 93.9~\hl{(${+}$\textbf{0.4})} \\     
        \midrule
        RX-50 & BN & 25.0 & 4.3 & \multicolumn{1}{l|}{77.6} & \multicolumn{1}{l}{93.7}\\
        RX-50 & BNET-$3$ & 25.1 & 4.3 & 78.0~\hl{(${+}$\textbf{0.4})} & 93.8~\hl{(${+}$\textbf{0.1})}\\ 
    \bottomrule
    \end{tabular}}
    \caption{Image classification results on ImageNet~\cite{deng2009imagenet} using ResNet~\cite{he2016deep} and ResNeXt~\cite{xie2017aggregated}.}
    \label{tab:resnet}
\end{table}
\subsubsection{Comparison on Lightweight Models}
\paragraph{MobileNetV2.} We apply BNET on an efficient architecture, \emph{i.e.}, MobilenetV2~\cite{Sandler2018MobileNetV2IR}. The Models are optimized by momentum SGD, with an initial learning rate of $0.05$ (annealed down to zero following a cosine schedule without restart), a momentum of 0.9, and a weight decay of $4\times10^{-5}$. They are trained by 8 GPUs with a batch size of 256 for a total of 150 epochs. 

Table~\ref{tab:mobile} shows that the Top-1 and Top-5 accuracies of MobileNetV2 with different widths using BN and BNET. We observe that BNET increases the Top-1 and Top-5 accuracies of MobileNetV2 (1.0) about 0.5\% and 0.6\%, respectively. Furthermore, the increase of the Top-1/Top-5 accuracy of MobileNetV2 becomes more significant as the width of the model is narrowed down.

\begin{table}[t]
\small
\resizebox{\linewidth}{!}{
    \centering
    \begin{tabular}{l|l|c|c|c|c}
    \toprule
        Model & Norm & \#Params & FLOPs & Top-1 & Top-5\\
        \midrule
        MBV2 (1.0) & BN & 3.5 & 300 & \multicolumn{1}{l|}{72.7} & \multicolumn{1}{l}{90.8}\\
        MBV2 (1.0) & BNET-$3$ & 3.5 & 330 & 73.2~\hl{(${+}$\textbf{0.5})} & 91.4~\hl{(${+}$\textbf{0.6})}\\
        \midrule
        MBV2 (0.5) & BN & 2.0 & 97 & \multicolumn{1}{l|}{65.1} & \multicolumn{1}{l}{85.9}\\
        MBV2 (0.5) & BNET-$3$ & 2.0 & 113 & 66.4~\hl{(${+}$\textbf{1.3})} & 86.7~\hl{(${+}$\textbf{0.8})} \\  
        \midrule
        MBV2 (0.25) & BN & 1.5 & 37 & \multicolumn{1}{l|}{53.6} & \multicolumn{1}{l}{76.9}\\
        MBV2 (0.25) & BNET-$3$ & 1.5 & 48 & 55.9~\hl{(${+}$\textbf{2.3})} & 78.6~\hl{(${+}$\textbf{1.7})}\\  
        \midrule
        Bi-real Net & BN & - &  162 & \multicolumn{1}{l|}{56.4} & \multicolumn{1}{l}{89.1} \\
        Bi-real Net & BNET-$3$ & - & 173 & 57.5~\hl{(${+}$\textbf{1.1})} & 89.4~\hl{(${+}$\textbf{0.3})}\\
        \bottomrule
    \end{tabular}}
    \caption{Image classification results on ImageNet~\cite{deng2009imagenet} using MobileNetV2~\cite{Sandler2018MobileNetV2IR} and Bi-real Net~\cite{liu2018bi}}
    \label{tab:mobile}
\end{table}

\paragraph{Bi-real Net.} Next we evaluate BNET on low-precision models. We adopt Bi-real-net~\cite{liu2018bi} as our basic model for its competitive results in binary models (both weights and features are binarized). We use the same training settings as the official PyTorch implementation\footnote{https://github.com/liuzechun/Bi-Real-net}. The Adam optimizer~\cite{kingma2014adam} is adopted with an initial learning rate 0.001 which linearly decays to 0 after 256 epochs. The batch size is set to 512 for 8 GPUs.

Table~\ref{tab:mobile} reports the results of Bi-real Net using BNET-$3$ and BN. With the help of BNET, the Top-1 and Top-5 accuracies of Bi-real net are significantly increased by 1.1\% and 0.3\%, respectively. It indicates that besides the full-precision models in Section~\ref{Sec:resnet}, BNET also benefits the training of low-precision models.
\subsubsection{Training with Auto-Augment}
To further demonstrate the out-standing fitting abilities of BNET, in this section, experiments of training with augmentations such as Auto-Augment~\cite{cubuk2019autoaugment} are presented. The training settings are the same as the settings in Section~\ref{Sec:resnet} except when the total epochs are 270. When the total epochs are set as 270, the learning rate is set to $0.1$ initially, and is multiplied by $0.1$ after every 80 epochs.

We report the results of ResNet-50 trained with Auto-Augment using BN and BNET in Table~\ref{tab:aug}. Two training epoch settings~(100 and 270) are provided for better comparisons. ResNet-50~(BN) is unable to fit the strong augmentation of Auto-augment with even a 0.2\% Top-1 accuracy drop compared to the results in Table~\ref{tab:resnet} when only trained 100 epochs. On the contrary, ResNet-50~(BNET-$3$) reveals better fitting abilities towards Auto-Augment with 0.4\% Top-1 accuracy gain over the model without Auto-Augment. Besides, ResNet-50~(BNET-$3$) outperforms ResNet-50~(BN) by a Top-1 error of 0.3\% when trained 270 epochs. Figure~\ref{fig.r50lossauto} and Figure~\ref{fig.r50accauto} illustrate the loss and accuracy curves of ResNet-50 which is trained 100 epochs with Auto-Augment. With the benefit of BNET, the loss drops more sharply compared with BN.
\begin{table}[t]
\small
\resizebox{\linewidth}{!}{
    \centering
    \begin{tabular}{l|l|c|c|c|c}
    \toprule
        Model & Norm &AA&  Epoch & Top-1&Top-5\\
        \midrule
        ResNet-50 & BN & \cmark & 100 & \multicolumn{1}{l|}{76.1} & \multicolumn{1}{l}{92.9}\\
        ResNet-50 & BNET-$3$ & \cmark & 100 & 77.2~\hl{(${+}$\textbf{1.1})} & 93.3~\hl{(${+}$\textbf{0.4})}\\
        \midrule
        ResNet-50 & BN & \cmark & 270 & \multicolumn{1}{l|}{77.5} & \multicolumn{1}{l}{93.7}\\
        ResNet-50 & BNET-$3$ & \cmark & 270 & 77.8~\hl{(${+}$\textbf{0.3})} & 94.0~\hl{(${+}$\textbf{0.3})}\\    
        \bottomrule
    \end{tabular}}
    \caption{Image classification results on ImageNet~\cite{deng2009imagenet} using ResNet~\cite{he2016deep}. AA denotes the Auto-Augment.}
    \label{tab:aug}
\end{table}
\subsection{Detection and Segmentation on MS-COCO}
We conduct object detection and instance segmentation experiments on MS-COCO~\cite{Lin2014MicrosoftCC} to evaluate the generalization performance of BNET on different tasks by using the models pre-trained on ImageNet as the backbones. The MS-COCO dataset has 80 object categories. We train the entire models on the trainval dataset, which is obtained by a standard pipeline that excludes $5\mathrm{K}$ images from the val set, merges the rest data into the $80\mathrm{K}$ train set and uses the minival set for testing. All the experiments on MS-COCO are implemented on the PyTorch~\cite{paszke2019pytorch} based MMDetection~\cite{chen2019mmdetection}.

\begin{figure}[!t]
	\renewcommand{\baselinestretch}{1.0}
	\centering
	\subfloat[Training loss]{
		\includegraphics[width=0.225\textwidth]{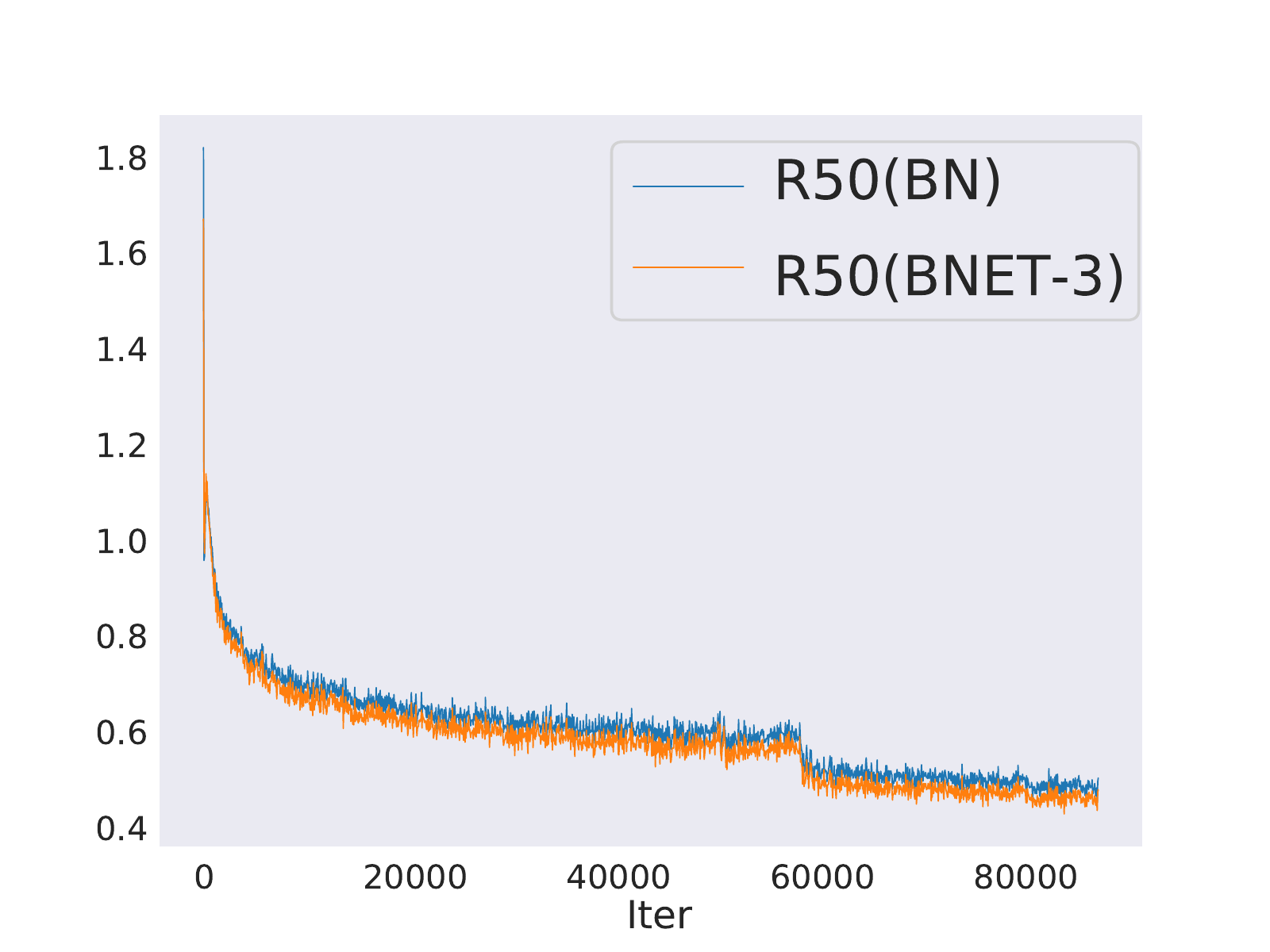}\label{fig.fasterloss}}
	\subfloat[Validation mAP]{
		\includegraphics[width=0.225\textwidth]{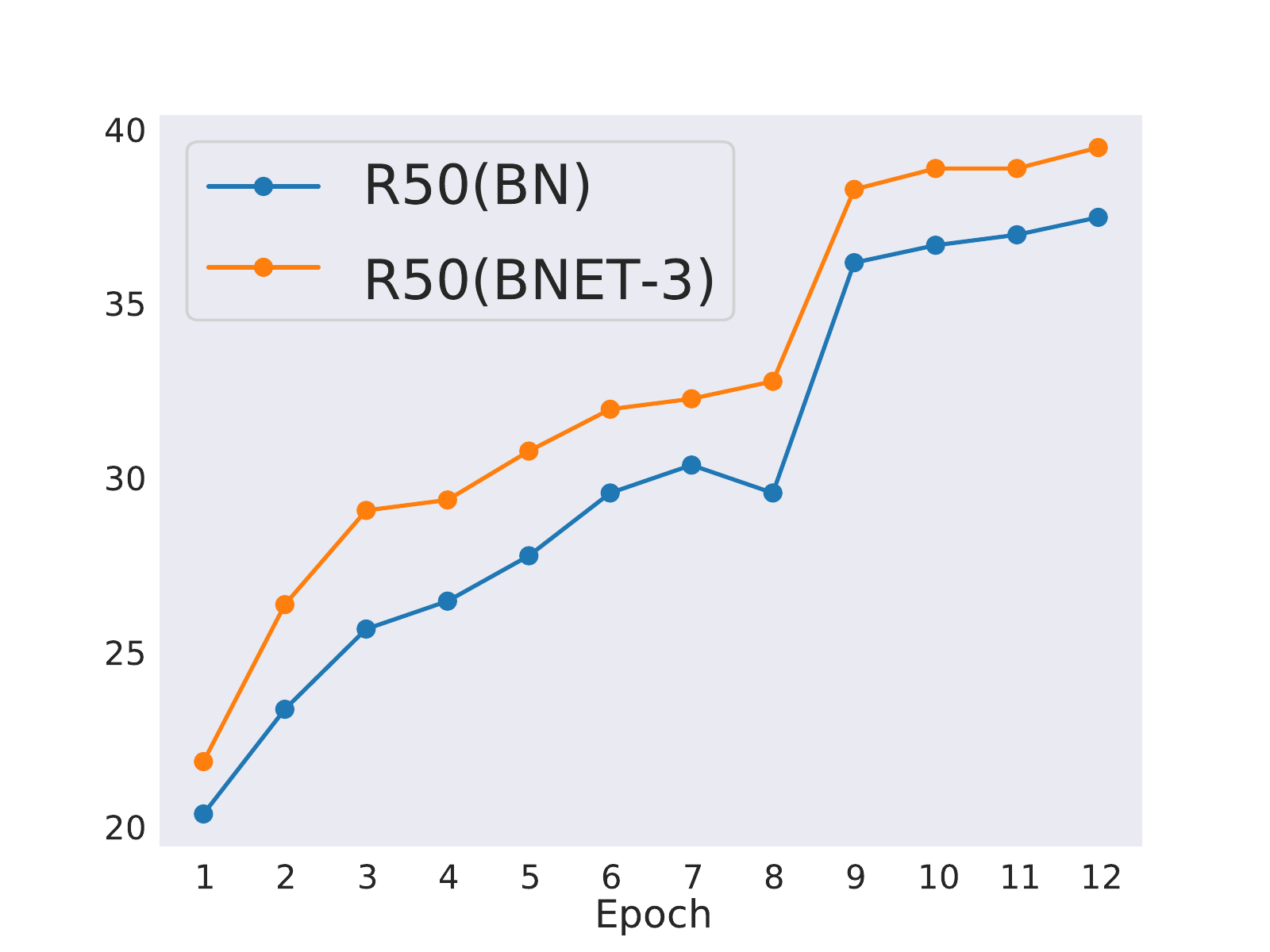}\label{fig.fastermap}}
	\caption{The training curves of Faster R-CNN on MS-COCO using ResNet-50 as the backbone.}\label{fig.faster}
\end{figure}
\subsubsection{Object Detection}
\begin{table*}[t]
\small
    \centering
    \resizebox{0.8\linewidth}{!}{
    \begin{tabular}{l|l|c|c|c|c|cc|ccc}
    \toprule
        Method & Backbone &\#Params&GFLOPs&FPS$^{\dagger}$& AP & AP$_{.5}$ & AP$_{.75}$ & AP$_{l}$ & AP$_{m}$ & AP$_{s}$ \\
        \midrule
        BN & ResNet-50 & 41.5 & 207.1 & 18.0 & \multicolumn{1}{l|}{37.5} & 58.4 & 40.6 & 21.0 & 41.1 & 48.3 \\
        BNET-$3$ & ResNet-50 & 41.6 & 208.1 & 17.1 & 39.5~\hl{(${+}$\textbf{2.0})} & 60.4 & 43.1 & 23.4 & 43.3 & 51.2 \\
        BNET-$5$ & ResNet-50 & 41.9 & 209.9 & 16.3 & 40.1~\hl{(${+}$\textbf{2.6})} & 61.3 & 43.8 & 23.7 & 43.5 & 52.5 \\        
        BNET-$7$ & ResNet-50 & 42.2 & 212.6 & 15.5 & 40.7~\hl{(${+}$\textbf{3.2})} & 62.0 & 44.2 & 24.2 & 44.0 & 52.5\\
        \midrule
        BN & ResNet-101 & 60.5 & 283.1 & 13.4 & \multicolumn{1}{l|}{39.4} & 60.1 & 43.1 & 22.4 & 43.7 & 51.1 \\
        BNET-$3$ & ResNet-101 & 60.8 & 284.8 & 13.0 & 40.7~\hl{(${+}$\textbf{1.3})} & 61.6 & 44.5 & 24.1 & 44.1 & 53.8 \\
        BNET-$5$ & ResNet-101 & 61.3 & 287.7 & 12.2 & 41.8~\hl{(${+}$\textbf{2.4})} & 62.6 & 45.5 & 24.4 & 45.7 & 54.5 \\
        \midrule
        BN & ResNeXt-101 & 60.2 & 286.9 & 11.7 & \multicolumn{1}{l|}{41.2} & 62.1 & 45.1 & 24.0 & 45.5 & 53.5 \\
        BNET-$5$ & ResNeXt-101 & 60.9 & 291.6 & 11.6 & 42.2~\hl{(${+}$\textbf{1.0})} & 63.3 & 46.1 & 24.8 & 46.3 & 55.2 \\
        \midrule
        BNET-$5$$^{\ddagger}$ & ResNet-50 & 41.9 & 209.9 & 16.3 & \multicolumn{1}{l|}{41.7} & 62.6 & 45.1 & 25.3 & 45.0 & 53.7  \\
        BNET-$5$$^{\ddagger}$ & ResNet-101 & 61.3 & 287.7 & 12.2 & \multicolumn{1}{l|}{43.1} & 63.6 & 46.4 & 27.1 & 47.0 & 55.7 \\
        \bottomrule
    \end{tabular}}
    \caption{Detection results on MS-COCO using Faster R-CNN~\cite{ren2015faster} and FPN~\cite{lin2017feature} with BN and BNET as normalization methods. $\dagger$ FPS is measured on a single Nvidia 1080Ti GPU and the batch size is set as 1. ${\ddagger}$ We also provide the results of the $3\times$ learning rate schedule with multi-scale training.}
    \label{tab:detection}
\end{table*}
\begin{table*}[h]
\small
\resizebox{\linewidth}{!}{
    \centering
    \begin{tabular}{l|l|ccc|ccc|ccc|ccc}
    \toprule
        Method & Backbone &  AP$^b$ & AP$^b_{.5}$ & AP$^b_{.75}$ & AP$^b_{l}$ & AP$^b_m$ & AP$^b_s$ & AP$^m$ & AP$^m_{.5}$ & AP$^m_{.75}$ & AP$^m_l$ & AP$^m_m$ & AP$^m_s$ \\
        \midrule
        BN & ResNet-50 & \multicolumn{1}{l}{38.2} & 58.8 & 41.4 & 21.9 & 40.9 & 49.5 & \multicolumn{1}{l}{34.7} & 55.7 & 37.2 & 18.3 & 37.4 & 47.2\\
        BNET-$3$ & ResNet-50 & \multicolumn{1}{l}{40.2~\hl{(${+}$\textbf{2.0})}} & 61.0 & 43.7 & 23.8 & 43.6 & 52.6 & 36.4~\hl{(${+}$\textbf{1.7})} & 58.0 & 38.7 & 19.9 & 39.5 & 49.4\\
        BNET-$5$ & ResNet-50 & \multicolumn{1}{l}{40.8~\hl{(${+}$\textbf{2.6})}}  & 61.8 & 44.4 & 23.9 & 44.3 & 53.5 & 36.7~\hl{(${+}$\textbf{2.0})} & 58.7 & 38.9 & 19.8 & 39.9 & 49.9\\
        \midrule
        BN & ResNet-101 & \multicolumn{1}{l}{40.0} & 60.5 & 44.0 & 22.6 & 44.0 & 52.6 & \multicolumn{1}{l}{36.1} & 57.5 & 38.6 & 18.8 & 39.7 & 49.5 \\
        BNET-$3$ & ResNet-101 & \multicolumn{1}{l}{42.2~\hl{(${+}$\textbf{2.2})}} & 63.2 & 46.0 & 25.5 & 46.1 & 54.8 & 37.8~\hl{(${+}$\textbf{1.7})} & 60.0 & 40.4 & 21.1 & 41.3 & 51.1\\
        BNET-$5$ & ResNet-101 & \multicolumn{1}{l}{42.5~\hl{(${+}$\textbf{2.5})}} & 63.1 & 46.7 & 25.2 & 46.5 & 55.7 & 37.9~\hl{(${+}$\textbf{1.8})} & 59.7 & 40.6 & 21.1 & 41.4 & 51.7\\
        \bottomrule
    \end{tabular}}
    \caption{Instance segmentation results on MS-COCO~\cite{Lin2014MicrosoftCC} using Mask R-CNN~\cite{he2017mask} and FPN~\cite{lin2017feature} with ResNet as backbone.
    }
    \label{tab:mask}
\end{table*}
We use the standard configuration of Faster R-CNN~\cite{ren2015faster} with FPN~\cite{lin2017feature} and ResNet/ResNeXt as the backbone architectures. The input image size is 1333$\times$800. We train the models on 8 GPUs with 2 images per GPU (effective mini-batch size of 16). The backbones of all models are pre-trained on ImageNet classification (Table~\ref{tab:resnet}) with the statistics of BN frozen. Following the 1$\times$ schedule in MMDetection, all models are trained for 12 epochs using synchronized SGD with a weight decay of 0.0001 and a momentum of 0.9. The learning rate is initialized to 0.02 and is decayed by a factor of 10 at the 9th and 11th epochs. Results of the 3$\times$ schedule are also provided.

Table~\ref{tab:detection} provides the results in terms of Average Precision~(AP) of Faster R-CNN trained with BN and BNET. There are several observations. First, BNET consistently outperforms BN on different backbone architectures. For example, ResNet-50~(BNET-$3$) has an increase of 2.0 AP comparing to the ResNet-50~(BN). Second, increasing the kernel size of BNET would further boost the detection scores as a bigger kernel size can bring more contextual information. Particularly, ResNet-101~(BNET-$7$) attains a 1.2\% gain in terms of AP over ResNet-101~(BNET-$3$). Third, in addition to precision, we also compare the computational cost including GFLOPs and FPS. The increased computational cost is negligible on all of the different backbone architectures. Detailed error analysis is presented in Appendix~C and the results of RetinaNet~\cite{lin2017focal} are shown in Appendix~B. The training curves of Faster-RCNN using ResNet-50 as the backbone are drawn in Figure~\ref{fig.faster}. The training loss of the model with BNET declines sharply and the AP score of BNET benefited model is consistently higher than the baseline model in each training epoch. 

\subsubsection{Instance Segmentation}
For the instance segmentation experiments, we use Mask R-CNN~\cite{he2017mask} with FPN~\cite{lin2017feature} as the basic framework and ResNet~\cite{he2016deep} as the backbone architecture. We follow the 1$\times$ schedule of MMDetection, which is the same as that in the detection experiments.

Comparison results of BNET and the baseline methods on instance segmentation are reported in Table~\ref{tab:mask}. Similar to the detection results of Faster R-CNN, we can also observe a great improvement of BNET compared with the original BN baseline model. The improvements become more significant when the kernel size of BNET is increased.

\subsection{Semantic Segmentation on Cityscapes}
Experiment results of semantic segmentation on Cityscapes~\cite{cordts2016cityscapes} are further presented to validate the effectiveness of BNET. The Cityscapes dataset contains 19 categories which includes 5,000 finely annotated images, 2,975 for training, 500 for validation, and 1525 for testing. We use the PSPNet~\cite{zhao2017pyramid} as the segmentation framework and the input image size is 512$\times$1024. For the training settings, we use the poly learning rate policy where the base learning rate is 0.01 and the
power is 0.9. We use a weight decay of 0.0005, and 8 GPUs with a batch size of 2 on each GPU to train $40\mathrm{K}$ iterations.

Table~\ref{tab:psp} summarizes the results of PSPNet with BNET and BN on the Cityscapes val set. With the similar model size and computational cost, BNET-$3$ achieves better performance, \textit{i.e.}, a 1.1\% gain over BN. 

\begin{table*}[!t]
   \centering
\resizebox{\linewidth}{!}{
   \begin{tabular}{l@{~}|@{~}c@{~~}c@{~~}c@{~~}c@{~~}c@{~~}c@{~~}c@{~~}c@{~~}c@{~~}c@{~~}c@{~~}c@{~~}c@{~~}c@{~~}c@{~~}c@{~~}c@{~~}c@{~~}c@{~}|c} \toprule
      Method                             & road          & sdwlk         & bldng         & wall          & fence         & pole          & light         & sign          & vgttn         & trrn          & sky           & person        & rider         & car           & truck         & bus           & train         & mcycl         & bcycl         & mIoU          \\
      \midrule
      BN & 97.6 & 81.5 & 91.6 & 42.0 & 56.3 & 63.3 & 70.5 & 78.2 & 92.3 & 65.4 & 94.3 & 81.2 & 60.7 & 94.8 & 75.0 & 88.0 & 78.3 & 62.6 & 76.8 & \multicolumn{1}{l}{76.3} \\
      BNET-$3$ & 98.2 & 84.9 & 92.4 & 53.7 & 57.4 & 64.3 & 70.9 & 78.5 & 92.5 & 63.5 & 94.3 & 81.2 & 60.4 & 95.1 & 77.0 & 87.8 & 77.7 & 63.8 & 76.8 & \multicolumn{1}{l}{77.4~\hl{(${+}$\textbf{1.1})}} \\
 \bottomrule
   \end{tabular}}
   \caption{\small Semantic segmentation results on Cityscapes~\cite{cordts2016cityscapes} using PSPNet~\cite{zhao2017pyramid} with ResNet-50~\cite{he2016deep} as backbone.}\label{tab:psp}
\end{table*}
\subsection{Action Recognition on UCF-101}
We further show the general applicability of BNET on the task of action recognition. Specifically, the experiments are conducted on the UCF-101~\cite{soomro2012ucf101} dataset. It contains $13\mathrm{K}$ videos which are annotated into 101 action classes. We compare BN and BNET on the first split of UCF-101.

We select TSN~\cite{wang2016temporal} as the base framework and use ResNet-50 as the backbone which is pre-trained on ImageNet~(Section~\ref{Sec:resnet}). For simplicity, we only use the RGB input. We utilize the SGD to learn the network parameters on 8 GPUs, where the batch size is set to 256 and momentum set to 0.9. The learning rate is initialized to 0.001, and decayed by a factor of 10 after every 30 epochs.

Table~\ref{tab:act} provides the results of TSN with different normalization methods. BNET-$3$ brings remarkable gains of 4.3\% and 1.5\% on Top-1 and Top-5 accuracies, respectively. However, BNET-$5$ is not able to give more improvements but more computational cost.

\begin{table}[!t]
\small
    \centering\resizebox{0.9\linewidth}{!}{
    \begin{tabular}{l|l|c|c|c}
    \toprule
        Method & Backbone & GFLOPs & Top-1 & Top-5\\
        \midrule
        BN      & ResNet-50 & 4.3 & \multicolumn{1}{l|}{79.8} & \multicolumn{1}{l}{96.0}\\
        \midrule
        BNET-$3$ & ResNet-50 & 4.4 &84.1~\hl{(${+}$\textbf{4.3})} & 97.5~\hl{(${+}$\textbf{1.5})} \\
        BNET-$5$ & ResNet-50  & 4.5 &84.2~\hl{(${+}$\textbf{4.4})} & 97.0~\hl{(${+}$\textbf{1.0})}\\
        \bottomrule
    \end{tabular}}
    \caption{Action recognition results of TSN~\cite{wang2016temporal} on UCF-101~(split $1$)~\cite{soomro2012ucf101} using ResNet~\cite{he2016deep} as backbone.}
    \label{tab:act}
\end{table}
\subsection{Ablation Study}
\begin{table}[!t]
\small   \centering
\resizebox{0.9\linewidth}{!}{
    \begin{tabular}{c|c|c|c|c}
    \toprule
         Linear Transformation & \#Params & GFLOPs& Top-1 & Top-5\\
        \midrule
          $1\times 1$& 25.6 & 4.1 & 76.3 & 93.0\\
        \midrule
          $3\times 3$& 25.7 & 4.2 & \textbf{76.8} & \textbf{93.1}\\
        \midrule
          $5\times 5$& 25.9 & 4.3 & 76.6 & 92.8\\
         \midrule
          $3\times 3$\ (dilation=2)& 25.7 & 4.2 & 76.5 & 92.9\\    
        \bottomrule
        \end{tabular}}
    \caption{Ablation studies about different  choices of linear transformations of BNET. Here, $1\times 1$ is equivalent to BN.}
    \label{tab:ablation2}
\end{table}

\paragraph{Comparisons of different linear transformations.} We first conduct experiments on different choices of linear transformation which may have different reception fields and embed diverse context information. We compare the performance of $\{1\times1, 3\times3, 5\times5, 3\times3\ (dilation=2)\}$ using ResNet-50 as the base model on ImageNet. 

Table~\ref{tab:ablation2} provides the results. From the results in the table, we can conclude that $3\times 3$ \textit{depth-wise convolution} is the best choice for BNET on ImageNet dataset. However, the performance of object detection and classification is not strictly consistent. As shown in Table~\ref{tab:detection} and Table~\ref{tab:mask}, $5\times 5$ \textit{depth-wise convolution} obtains better performance on the MS-COCO even though the pre-trained performance is worse.  This phenomenon indicates that a task like Object Detection prefers a larger kernel size $k$ as a larger kernel-size $k$ in BNET-$k$ can produce more contextual information and a broader reception field.

\paragraph{Discussions about the plug-in position.} We conduct experiments about where to add BNET to offer a better trade-off between accuracy and computational cost. There exist three BN layers in a residual bottleneck block which is denoted as A, B and C, respectively. A and C are the BN layers after the $1\times 1$ convolution and B is the BN after the $3\times 3$. Experiment results are shown in Table~\ref{tab:ablation1}. We conclude that replacing the BN layer in C is the best choice this may because C has wider input channels than A and C. Further, we consider the case that replacing all the BN layers in A, B and C by BNET. However, it can not provide extra improvement which means that configuration C has provided enough contextual information.

\begin{table}[!t]
\small
    \centering\resizebox{0.9\linewidth}{!}{
    \begin{tabular}{l|c|c|c|c|c|c|c}
    \toprule
        Model & A & B & C & \#Params & GFLOPs & Top-1&Top-5\\
        \midrule
        ResNet-50 & \cmark &  &  & 25.6 & 4.1 & 76.5&93.0\\
        \midrule
        ResNet-50 &  & \cmark &  & 25.6 & 4.1 & 76.3 & 92.9\\
        \midrule
        ResNet-50 &  &  & \cmark & 25.7 & 4.2 & \textbf{76.8} & \textbf{93.1}\\
         \midrule
        ResNet-50 & \cmark & \cmark & \cmark & 25.7 & 4.2 & 76.7 & 93.0\\    
        \bottomrule
    \end{tabular}}
    \caption{The positions of BNET in the residual Bottleneck}
    \label{tab:ablation1}
\end{table}

\begin{table}[!t]
\small   
\centering
\resizebox{0.9\linewidth}{!}{
    \begin{tabular}{l|c|c|c|c}
    \toprule
         Model & \#Params & GFLOPs& Top-1 & Top-5\\
        \midrule
         ResNet-50 $+$ BN & 25.6 & 4.1 & 76.3 & 93.0\\
        \midrule
         ResNet-50 $+$ BNET & 25.7 & 4.2 & \textbf{76.8} & \textbf{93.1}\\
        \midrule
         ResNet-50 $+$ BN $+$ Conv & 25.7 & 4.2 & 75.7 & 92.6\\
        \bottomrule
    \end{tabular}}
    \caption{Compared with adding a convolution layer by using ResNet-50 on ImageNet.}
    \label{tab:ablation3}
\end{table}
\begin{table}[!t]
\small
    \centering
    \resizebox{0.9\linewidth}{!}{
    \begin{tabular}{l|c|c|c}
    \toprule
         Model & Enhancement& Top-1 & Top-5\\
        \midrule
         ResNet-50 $+$ BN & \xmark & 76.3 & 93.0\\
         ResNet-50 $+$ BN & \cmark & 76.8 & 93.1\\
        \midrule
         ResNet-50 $+$ GN & \xmark & 75.7 & 92.7\\
         ResNet-50 $+$ GN & \cmark & 76.0 & 92.8\\
        \midrule
         ResNet-50 $+$ SN & \xmark & 76.9 & 93.0\\
         ResNet-50 $+$ SN & \cmark & 77.2 & 93.1\\
        \bottomrule
    \end{tabular}}
    \caption{Applications on different normalization layers by using ResNet-50 on ImageNet.}
    \label{tab:differentnorm}
    \vspace{-1em}
\end{table}

\begin{figure*}[!t]
\centering
\includegraphics[width=0.85\textwidth]{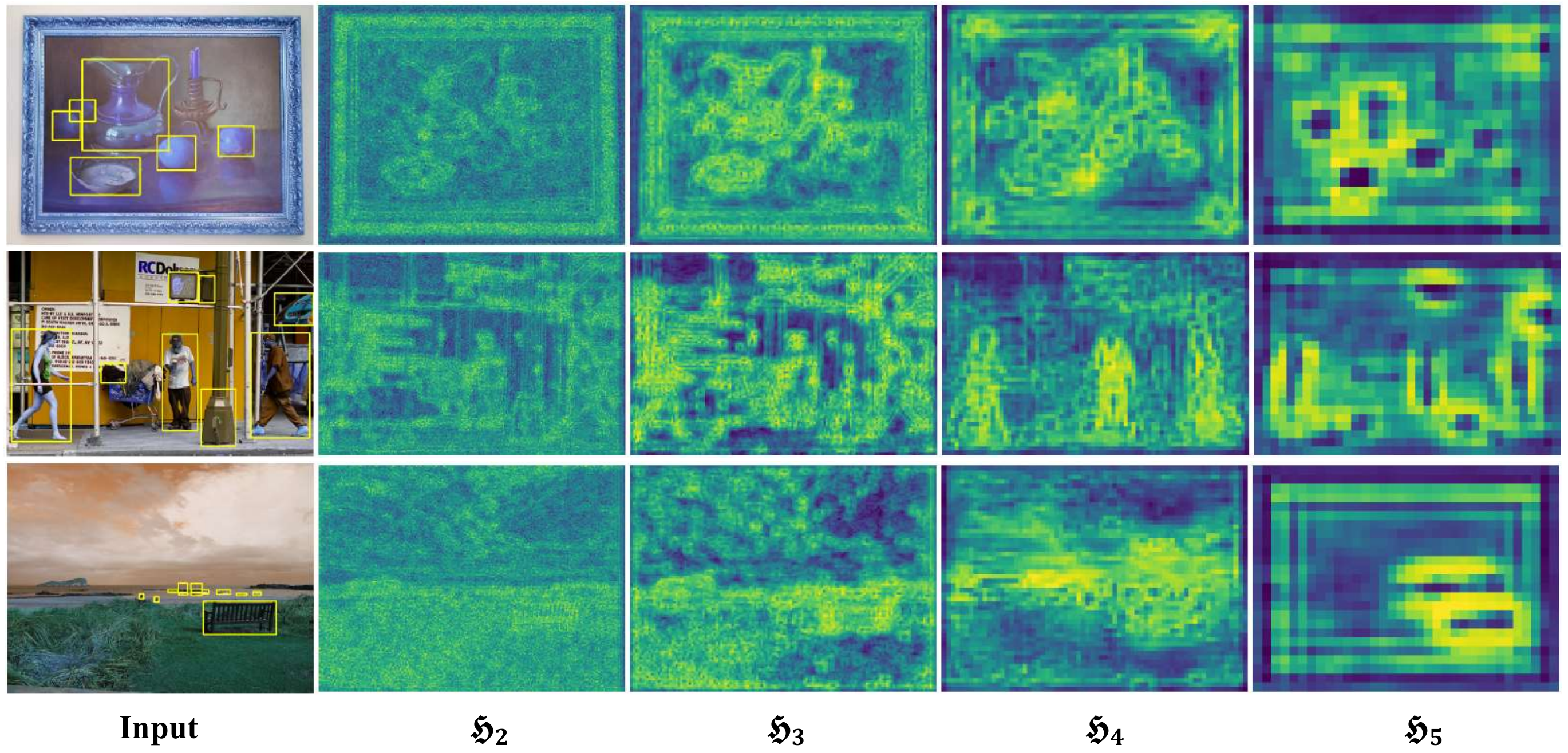}
\caption{The visualization of enhancement heat-maps of BNET-$3$. The first row is the input image from MS-COCO with bounding box and $\{\mathfrak{H}_2,\mathfrak{H}_3,\mathfrak{H}_4,\mathfrak{H}_5\}$ are enhancement heat-maps of BNET in each stage’s last residual block which corresponds to the responses fed into the feature pyramid network. }
\vspace{-1em}
\label{fig:heatmap}
\end{figure*}

\paragraph{Comparison to adding a convolution layer.} At first glance, the proposed BNET looks very simple. An immediate concern arises: what is the difference with simply adding a convolution layer in the residual block? To answer this question, we perform additional experiments and compare the results in Table~\ref{tab:ablation3}. We add a ``ReLU" layer followed by a ``$3\times 3$ \textit{depth-wise convolution}" layer in the residual bottleneck. The results show that simply adding a convolution layer in the bottleneck is unable to boost performance, on the contrary, the Top-1 and Top-5 accuracy even drops by 0.6\% and 0.4\%, respectively.


\paragraph{Adding contextual information to other normalization layers.} In addition to BN, many normalization methods use the same linear recovery as BN, such as IN~\cite{ulyanov2016instance}, SN~\cite{luo2018differentiable} and GN~\cite{wu2018group}. Can contextual information boost their performance? To answer this question, we enhance the linear transformations of GN and SN which is similar to BNET-$3$. Table~\ref{tab:differentnorm} provides the experiment results of ResNet-50 on ImageNet. In addition to BN, both of GN and SN can benefit from the contextual information.

\subsection{What Was Enhanced by BNET?}\label{Sec:Disc}

To answer the question, we delve deep into BNET by observing the behavior of the linear transformation. We use the Faster R-CNN model (equipped with BNET-$3$) trained for MS-COCO object detection. We choose four layers, denoted by $\mathfrak{H}_2$--$\mathfrak{H}_5$, that correspond to the responses fed into the feature pyramid network. For each layer, we investigate the input-output pairs (before and after the linear transformation of BNET). Note that in the vanilla BN, these pairs follow a simple linear correlation, but using BNET, we can regress a linear function upon these pairs and observe which outputs are larger than the regressed value. We record the corresponding positions for the enhanced outputs and sum up them across all the channels, and obtain the enhancement heatmap. We show these heatmaps in Figure~\ref{fig:heatmap}.

One can observe that mostly, the enhanced positions correspond to meaningful contents, which can be interesting textures in low-level layers (\textit{e.g.}, $\mathfrak{H}_2$) and objects in high-level layers (\textit{e.g.}, $\mathfrak{H}_5$). These results indicate that BNET has the ability to enhance important regions. We conjecture that this property contributes to the improved recognition accuracy as well as the faster convergence of BNET.


\section{Conclusion}\label{Sec:Con}

In this paper, we propose BNET, a variant of BN that enhances the linear transformation module by referring to the neighboring responses of each neuron. BNET is easily implemented, requires little extra computational costs, and achieves consistent accuracy gain in various visual recognition tasks. We demonstrate that designing a more sophisticated module to balance normalization and flexibility leads to better performance, and believe that this research topic has higher potentials in the future.

{\small
\bibliographystyle{ieee_fullname}
\bibliography{egbib}
}


    \newcommand{\uparr}{\rotatebox[origin=c]{90}{\ding{213}}}
    \newcommand{\downarr}{\rotatebox[origin=c]{270}{\ding{213}}}
    
    \newcommand{\dmAP}[1]{\text{E}_\texttt{#1}{\text{\downarr}}}
    \newcommand{\spacer}{\phantom{aa}}
    \newcommand{\halfspacer}{\phantom{a}}

\begin{table*}[h]
\small
    \centering
    \resizebox{0.8\linewidth}{!}{
    \begin{tabular}{l|l|c|c|c|c|cc|ccc}
    \toprule
        Method & Backbone &\#Params&GFLOPs&FPS$^{\dagger}$& AP & AP$_{.5}$ & AP$_{.75}$ & AP$_{l}$ & AP$_{m}$ & AP$_{s}$ \\
        \midrule
        BN & ResNet-101 & 56.7 & 315.4 & 12.8 & \multicolumn{1}{l|}{38.5} & 57.6 & 41.0 & 21.7 & 42.8 & 50.4 \\
        BNET-$5$ & ResNet-101 & 57.5 & 320.0 & 11.4 & 40.7~\hl{(${+}$\textbf{2.2})} & 60.5 & 43.3 & 23.0 & 44.6 & 54.3 \\        
        \bottomrule
    \end{tabular}}
    \caption{Detection results on MS-COCO using RetinaNet~\cite{lin2017focal} and FPN~\cite{lin2017feature} with BN and BNET as normalization methods. \\$\dagger$ FPS is measured on a single Nvidia 1080Ti GPU and the batch size is set as 1.}
    \label{tab:retina}
\end{table*}

\begin{table*}[h]
\small
    \centering
        
    \resizebox{0.8\linewidth}{!}{
    \begin{tabular}{l r r r r rrrr rr rr}
    \toprule
                  Method && $\text{AP}{\text{\uparr}}$ &$\text{AP}_{50}{\text{\uparr}}$ && $\dmAP{cls}$ & $\dmAP{loc}$ & $\dmAP{bkg}$ & $\dmAP{miss}$ &&& $\dmAP{FP}$ & $\dmAP{FN}$ \\
        \midrule
 ResNet-50~(BN)       & &            37.5 & 58.4 & &         3.9 &          7.1 &        3.8 &           7.9 &&&        17.6 &        16.0 \\
 \midrule
 ResNet-50~(BNET-$3$) & &            39.5 & 60.4 & &         3.5 &          6.9 &        3.8 &           7.8 &&&        17.1 &        15.3 \\
 Improvement          & &           ~\hl{\textbf{$+2.0$}} & ~\hl{\textbf{$+2.2$}} & &      ~\hl{\textbf{$-0.4$}} &       ~\hl{\textbf{$-0.2$}} & ~\hl{\textbf{$-0.0$}} &        ~\hl{\textbf{$-0.1$}} &&&      ~\hl{\textbf{$-0.5$}} &      ~\hl{\textbf{$-0.7$}} \\
 \midrule
 ResNet-50~(BNET-$5$) & &            40.1 & 61.3 & &          3.2 &          7.1 &        3.8 &           7.7 &&&        17.1 &        15.1 \\
 Improvement          & &           ~\hl{\textbf{$+2.6$}} & ~\hl{\textbf{$+2.9$}} & &       ~\hl{\textbf{$-0.7$}} &       ~\hl{\textbf{$-0.0$}} & ~\hl{\textbf{$-0.0$}} &        ~\hl{\textbf{$-0.2$}} &&&      ~\hl{\textbf{$-0.5$}} &      ~\hl{\textbf{$-0.9$}} \\
 \midrule
 ResNet-50~(BNET-$7$) & &            40.7 & 62.0 &  &          3.2 &          6.9 &        3.7 &           8.0 &&&        16.2 &        15.6 \\
 Improvement          & &           ~\hl{\textbf{$+3.2$}} & ~\hl{\textbf{$+3.6$}} & &       ~\hl{\textbf{$-0.7$}} &       ~\hl{\textbf{$-0.2$}} & ~\hl{\textbf{$-0.1$}} &        ~\hl{\textbf{$+0.1$}} &&&      ~\hl{\textbf{$-1.4$}} &      ~\hl{\textbf{$-0.4$}} \\
\bottomrule
    \end{tabular}}
    \caption{Error analysis on MS-COCO using Faster R-CNN~\cite{ren2015faster} and FPN~\cite{lin2017feature} with BN and BNET as normalization methods. The error analysis is performed by means of TIDE~\cite{bolya2020tide}.}
    \label{tab:tide}
\end{table*}

\begin{table}[!t]
\small
\setlength{\tabcolsep}{0.12cm}
    \centering
    \begin{tabular}{l|l|l|c}
    \toprule
        Model & Backbone & Norm& mIOU\\
        \midrule
        DeeplabV3 & ResNet-101 & BN       & \multicolumn{1}{l}{76.5}\\
        DeeplabV3 & ResNet-101 & BNET-$3$ & 77.5~\hl{(${+}$\textbf{1.0})}\\
        \bottomrule
    \end{tabular}
    \caption{Semantic segmentation results of DeeplabV3~\cite{chen2017rethinking} using ResNet101 as the backbone architecture on PASCAL VOC 2012~\cite{everingham2010pascal}}
    \label{tab:pascal}
\end{table}

\begin{appendix}
\section{Visualizations of the ResNet-50~(BNET)}
\begin{figure*}[!t]
\centering
\includegraphics[width=0.9\textwidth]{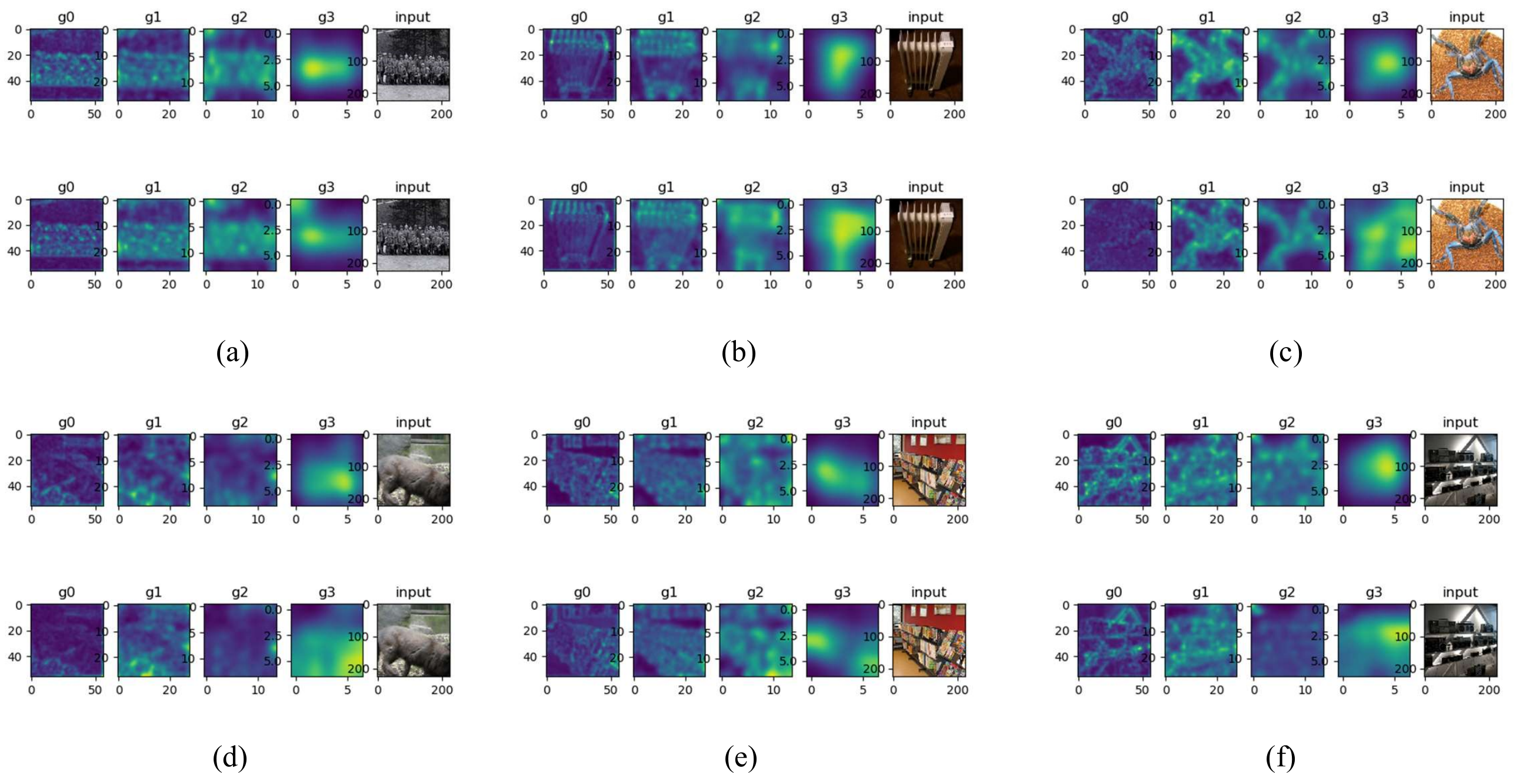}
\caption{Attention maps of different stages in ResNet-50~(BNET-$3$). Six input images from ImageNet are fed into ResNet50~(BN) and ResNet~(BNET-$3$). In each sub-figure, the first row is the attentions maps of ResNet~(BNET-$3$) and the second row belong to ResNet50~(BN). }
\label{fig:attention}
\end{figure*}

Figure~\ref{fig:attention} depicts the attention maps of ResNet-50~(BN) and ResNet-50~(BNET). $g_0,g_1,g_2,g_3$ are the attention maps of the last residual block in each stage of ResNet-50. One can observe that, the attention maps of ResNet~(BNET) are more focused on the target object (\emph{e.g.}, $g_3$) when compared with the attention maps of ResNet-50~(BN).
\section{RetinaNet Results on MS-COCO}
We use RetinaNet~\cite{lin2017focal} as the basic framework using ResNet-101 as the backbone architecture. Experiments are performed using MMdetection~\cite{chen2019mmdetection}. The backbones are pre-trained on ImageNet classification with the statistics of BN frozen. Following the 1$\times$ schedule in MMDetection, all models are trained for 12 epochs using synchronized SGD with a weight decay of 0.0001 and a momentum of 0.9. The learning rate is initialized to 0.02 and is decayed by a factor of 10 at the 9th and 11th epochs.

Table~\ref{tab:retina} provides the results of RetinaNet with BNET and BN as the normalization methods. The model using BNET obtains a gain of 2.2\% over the model using BN.

\section{Error Analysis on MS-COCO}
We provide the analysis using the tool TIDE~\cite{bolya2020tide} on MS-COCO. We adopt Faster R-CNN and use four kinds of ResNet-50 backbone architectures, \emph{i.e.}, ResNet-50~(BN), ResNet-50~(BNET-$3$), ResNet-50~(BNET-$5$), and ResNet-50~(BNET-$7$).

\begin{figure*}[h]
	\renewcommand{\baselinestretch}{1.0}
	\centering
	\subfloat[BN]{
		\includegraphics[width=0.204\textwidth]{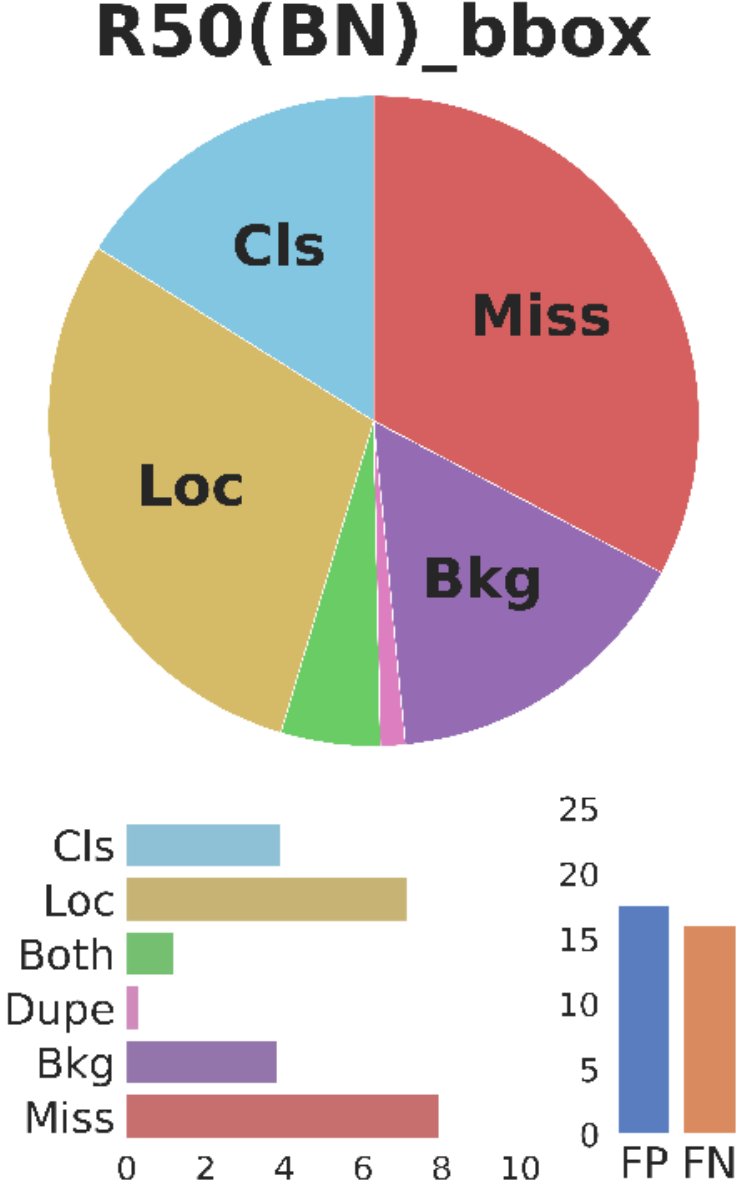}\label{fig.tidebn}}\hspace{5mm}
	\subfloat[BNET-$3$]{
		\includegraphics[width=0.2\textwidth]{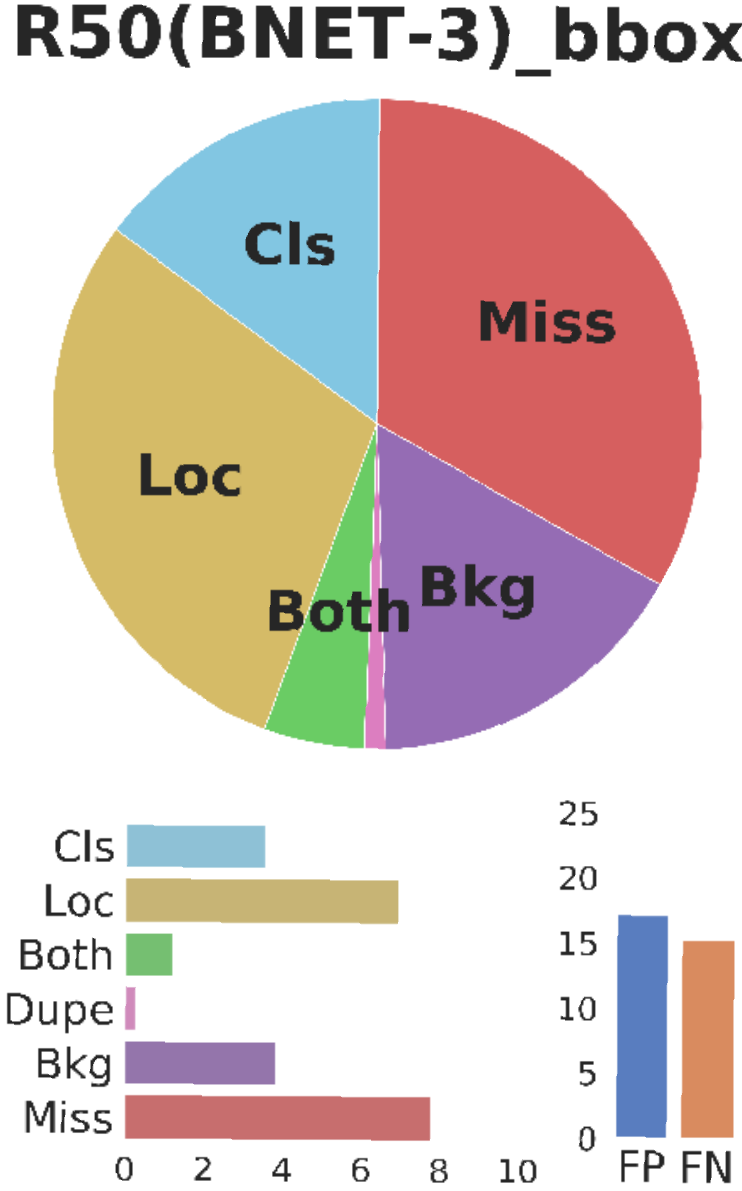}\label{fig.tidebnet3}}\hspace{5mm}
	\subfloat[BNET-$5$]{
		\includegraphics[width=0.2\textwidth]{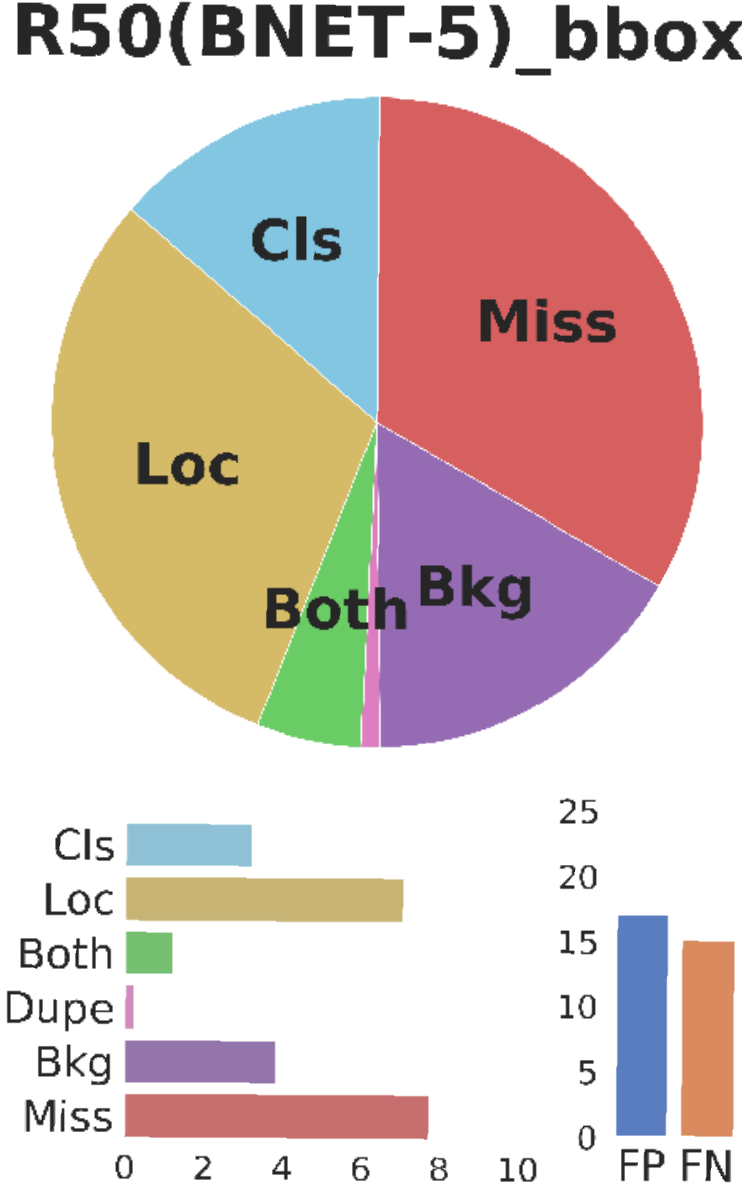}\label{fig.tidebnet5}}\hspace{5mm}
	\subfloat[BNET-$7$]{
		\includegraphics[width=0.2\textwidth]{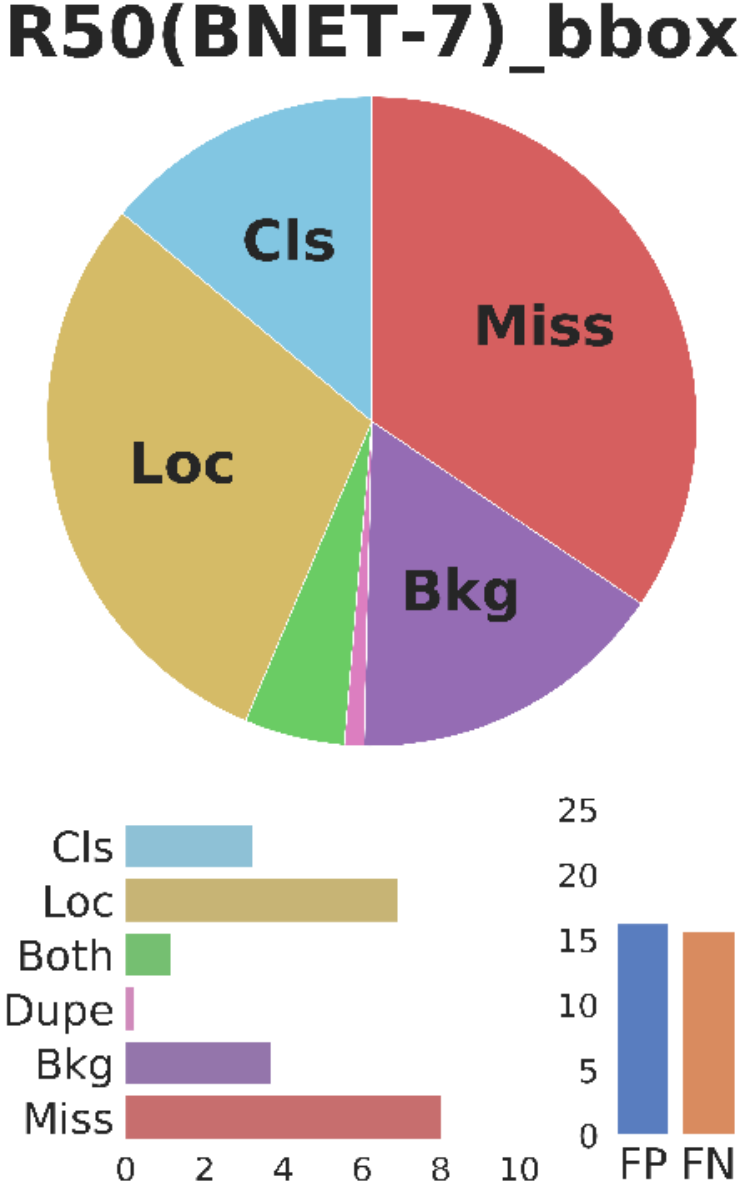}\label{fig.tidebnet7}}
	\caption{The relative contributions of different errors in Faster R-CNN~\cite{ren2015faster} with FPN~\cite{lin2017feature} using ResNet-50~(BN), ResNet-50~(BNET-$3$), ResNet-50~(BNET-$5$), and ResNet-50~(BNET-$7$) as the backbone architectures.}\label{fig.TIDE}
\end{figure*}

Table~\ref{tab:tide} provides the error analysis of BNET and BN based models on MS-COCO. Follow \cite{bolya2020tide}, we calculate four kinds of sub-classified errors~(\emph{e.g.}, Classification Error $(E_{cls)}$, Localization Error $(E_{loc)}$, Background Error $(E_{bkg)}$, and Missed GT Error $(E_{miss)}$), False Positive Error and False Negative Error. All the errors are calculated by the $\Delta AP$ proposed in \cite{bolya2020tide}. All the three models with BNET decrease the classification error greatly. The localization errors of BNET-$3$ and BNET-$7$ have also decreased. We also plot the relative contribution of each error in Figure~\ref{fig.TIDE}.

\section{Segmentation Results on VOC2012}
In this section, we provide experiment results of semantic segmentation on PASCAL VOC 2012~\cite{everingham2010pascal} to compare BNET with BN baseline. The PPASCAL VOC dataset contains 20 foreground object classes and one background class. We augment the original dataset with the extra annotations which contains 10,582 (train aug) training images. We use the DeeplabV3~\cite{chen2017deeplab} as the segmentation framework.

Table~\ref{tab:pascal} summarizes the results of DeeplabV3 with BNET and BN on the PASCAL VOC 2012 set. With similar model size and computational cost, BNET-$3$ achieves better performance: a 1.0\% gain over BN. 

\end{appendix}

\end{document}